\documentclass[5p,times,authoryear]{elsarticle}
\usepackage[T1]{fontenc}
\usepackage[utf8]{inputenc}
\usepackage{amsmath,amssymb,bm}
\usepackage{graphicx,booktabs,multirow,threeparttable}
\usepackage{adjustbox}
\usepackage{microtype}
\usepackage{dblfloatfix}
\usepackage{placeins}
\usepackage{caption}
\usepackage{xcolor,url}
\usepackage{hyperref}
\biboptions{round}
\journal{Expert Systems with Applications}
\date{}
\hypersetup{colorlinks=true,linkcolor=black,citecolor=black,urlcolor=black,pdfauthor={}}
\AtBeginDocument{\hypersetup{linkcolor=black,citecolor=black,urlcolor=black}}
\begin{document}
\begin{frontmatter}
	\title{Facial Action Unit Detection Based on Multi-task Learning Strategy for Unlabeled Facial Images in the Wild}
	
	\author[1]{Ziqiao Shang}
	\ead{M202172456@hust.edu.cn}
	
	\author[2]{Bin Liu\corref{cor1}}
	\ead{binliu@swjtu.edu.cn}
	
	\address[1]{School of Electronic Information and Communications, Huazhong University of Science and Technology (HUST), Wuhan 430074, China}
	
	\address[2]{School of Computing and Artificial Intelligence, Southwest Jiaotong University}
	
	\cortext[cor1]{Corresponding author}
	
	\begin{abstract}
		Facial Action Unit (AU) detection often relies on highly-cost accurate labeling or inaccurate pseudo labeling techniques in recent years. How to introduce large amounts of unlabeled facial images in the wild into supervised AU detection frameworks has become a challenging problem. Additionally, nearly every type of AUs has the problem of unbalanced positive and negative samples. Inspired by other multi-task learning frameworks, we first propose a multi-task learning strategy boosting AU detection in the wild through jointing facial landmark detection and AU domain separation and reconstruction. Our introduced dual domains facial landmark detection framework can solve the lack of accurate facial landmark coordinates during the AU domain separation and reconstruction training process, while the parameters of homostructural facial extraction modules from these two similar facial tasks are shared. Moreover, we propose a pixel-level feature alignment scheme to maintain the consistency of features obtained from two separation and reconstruction processes. Furthermore, a weighted asymmetric loss is proposed to change the contribution of positive and negative samples of each type of AUs to model parameters updating. Experimental results on three widely used benchmarks demonstrate our superiority to most state-of-the-art methods for AU detection.
	\end{abstract}
	
\end{frontmatter}

\section{Introduction} \label{Introduction}
Facial Action Coding System (FACS)~\citep{rosenberg:et.al:2020} is a widely used method for coding facial expressions. It has proposed a set of 44 facial action units (AUs) that establish a connection between facial muscle movements and facial expressions~\citep{dahmane:et.al:2014:TMM}, allowing for their simulation. In recent years, AU detection~\citep{zhao:et.al:2016:CVPR,corneanu:et.al:2018:ECCV,li:et.al:2018:TPAMI,ma:et.al:2019:Neurocomputing} has become a technique that uses computer vision algorithms to automatically detect and track the AUs in videos or images, which is now promising applications in diverse fields, including human-computer interaction~\citep{bauer:et.al:2008:IJHR}, emotion analysis~\citep{mcduff:et.al:2013:FG,szirtes:et.al:2013:ER}, and car-driving monitoring~\citep{vicente:et.al:2015:TITS}.

Recently, due to the high labor cost to obtain a large number of high-precision AU labels for images, self-supervised learning has garnered significant attention for AU detection~\citep{wiles:et.al:2018:BMVC-a,wiles:et.al:2018:BMVC-b,li:et.al:2019:CVPR,lu:et.al:2020:BMVC,sun:et.al:2021:FG,li:et.al:2022:TPAMI,li:et.al:2023:TIP}. These methods do not rely on costly manual labeling. However, most self-supervised methods proposed so far mainly concentrate on images in laboratory scene, which have low generalization performance in the wild. Moreover, some semi-supervised and weakly-supervised methods~\citep{wang:et.al:2017:arXiv,zhao:et.al:2018:CVPR,niu:et.al:2019:NIPS,yan:et.al:2023:TMM,yang:et.al:2023:ICONIP,liu:et.al:2023:ICASSP} are introduced into AU detection, which combine labeled and unlabeled images in training process, and generate pseudo labels through reasoning on unlabeled images. However, the pseudo labels in the wild are also automatically labeled by AU detection frameworks trained in laboratory scene, which can not solve the problem of the large domain distribution difference between annotated images and training images. Comparing images captured in controlled laboratory settings to those taken in the wild, there are notable disparities in various aspects. These disparities encompass variations in facial expressions, poses, ages, lighting conditions, accessories, occlusions, backgrounds, and the overall quality of the images. Therefore, how to make AU detection have strong generalization in the wild is a challenging task.

To solve the problem of imprecise pseudo labeling in the wild, we introduce a novel facial task called AU domain feature separation and reconstruction, which can transfer AU knowledge from a source domain in laboratory scene with accurate AU labels to a target domain in the wild without AU labels. One kind of schemes is to learn features that are invariant across different domains~\citep{ganin:et.al:2016,tzeng:et.al:2017:CVPR}. However, these approaches may lead to the loss of AU-related information due to the interdependence between AUs and head poses. Another kind of schemes involves transforming source-domain images to resemble the style of the target domain~\citep{lee:et.al:2018:ECCV,zheng:et.al:2018:ECCV}. Nonetheless, these approaches alone may not adequately address other forms of domain shift caused by factors like variations of head poses or facial occlusions. Additionally,~\citet{shao:et.al:2022:TAC} propose to mix facial features from source domain and backgrounds from target domain. This method preserve facial features by accurate facial landmark ground truths. However, it is very unpractical to obtain a large amount of accurate facial landmark ground truths during the training process, especially in target domain.

Due to the similarity in feature extraction for facial tasks, multi-task learning training strategies are widely applied in AU detection~\citep{jyoti:et.al:2018:DICTA,tallec:et.al:2022:TAC}. These strategies provide more flexible, efficient, and robust ways to learn facial features among different facial tasks, which also help improve the generalization performance of the models extracting facial features. By training on multiple facial tasks, the models can learn richer, generic features that make it better able to adapt to new facial tasks or previously unseen facial images. In previous work~\citep{shao:et.al:2021:IJCV,wang:et.al:2023:TAC}, facial landmark detection is the most common task combined with AU detection due to the fact that AU centers can be located by the results of facial landmark detection, so as to realize the regional learning of AUs. However, many AU domain separation and reconstruction frameworks also need to use accurate facial landmark coordinates and AU labels, and there is no multi-task learning framework to combine these three tasks together.

$\textbf{Our proposed solutions.}$ Inspired by the above issues, we propose a multi-task learning strategy boosting AU detection in the wild through jointing facial landmark detection and AU domain separation and reconstruction. To make AU detection have strong generalization in the wild, the introduced AU domain separation and reconstruction framework can realize the combination of AU-related features from source domain and background features from target domain. Different from utilizing inaccurate pseudo labels, AU labels from source domain are accurate, and it can also generalize the performance in the wild because of the background features from target domain during the training process. This way not only addresses complex labeling with high labor cost, but also solve the problem of additional domain shifts arising from head pose variations and occlusions. 

\begin{figure}[tbp]
	\centering
	\includegraphics[width=0.99\linewidth]{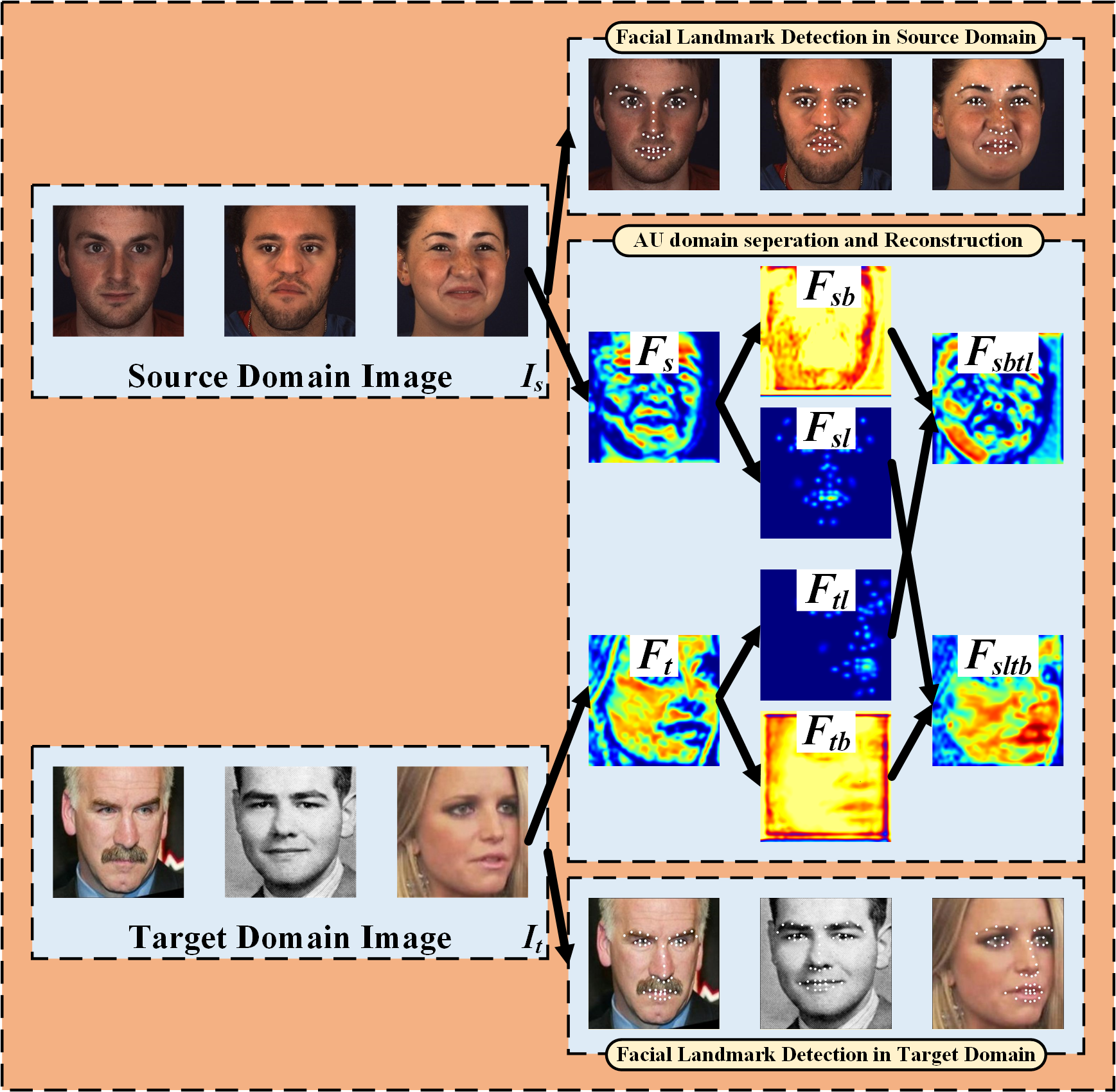}
	\caption{Illustration of jointly learning facial landmark detection with AU domain seperation and reconstruction.}
	\label{fig:1.png}
\end{figure}

Furthermore, the introduced dual domains facial landmark detection framework can enable the framework to use accurate facial landmark coordinates and take advantage of the strong correlation among different facial tasks with similar features. While precise AU ground truths are not accessible for the target domain, our proposed facial landmark framework can generate accurate facial landmark labels highly relevant to AUs. These labels are utilized to predefine AU centers. The facial landmark ground truths for the target domain are readily obtainable through contemporary landmark detection methods~\citep{king:et.al:2012:access,zhu:et.al:2016:CVPR,simon:et.al:2017:CVPR,wu:et.al:2018:CVPR} exhibiting high accuracy comparable to manual labeling. More importantly, our proposed multi-task learning strategy can jointly learn above two facial tasks by sharing the parameters of facial extraction modules with the same structure, which also promotes the precision of AU detection. The learning process of two above facial tasks is illustrated in Fig.~\ref{fig:1.png}.

In addition, inspired by some self-supervised representation learning frameworks~\citep{li:et.al:2019:CVPR,chang:et.al:2022:CVPR}, we use a cross-cyclically model for feature separation and reconstruction instead of a model separated and reconstructed features only once. The advantage of the cross-cyclically mechanism is to improve the quality of facial feature reconstruction by requiring the regenerated features of both source and target domains from the second separated and reconstructed branch to be consistent with the original features from the first separated and reconstructed branch through unsupervised reconstruction loss. Therefore, to maintain the consistency of features obtained from two separation and reconstruction processes, we propose a pixel-level feature alignment scheme, which is realized by projectors with simple structure and an improved self-reconstruction loss. Compared to feature reconstruction processes of other self-supervised frameworks~\citep{shao:et.al:2022:TAC,bhattacharya:et.al:2020:ECCV}, our framework uses projectors to magnify the receptive field of feature pixel elements to achieve better alignment, and ensures the integrity of the whole reconstruction process by adding four intermediate supervisors, while also increasing the interpretability of the framework. Apart from that, to solve the serious imbalance problem of positive and negative samples of each type of AUs, we propose a weighted asymmetric loss to respectively change the contribution of positive and negative samples to model parameters updating. The main contributions of our work are summarized as follows:
\begin{itemize}
	\item We propose a multi-task learning strategy boosting AU detection through jointing facial landmark detection and AU domain separation and reconstruction. To our knowledge, this is the first work to jointly learn these two facial tasks for AU detection.
	\item We propose a pixel-level feature alignment scheme. The simple projectors are developed for magnifying the receptive field of feature pixel elements and an improved self-supervised loss adding four intermediate supervisors is proposed for maintaining the consistency of features obtained from two separation and reconstruction processes.
	\item We propose a weighted asymmetric loss for respectively changing the contribution of positive and negative samples of each type of AUs to model parameters updating. 
	\item Extensive experiments on three widely-used datasets demonstrate that our method soundly outperforms most state-of-the-art techniques. 
\end{itemize} 
\FloatBarrier
\section{Related Work}
\label{Sec:Related Work}
	We mainly review three techniques that are most relevant to our work, including self-supervised and semi-supervised AU detection, AU domain separation and reconstruction and multi-task learning frameworks for AU detection.
	
\subsection{Self-supervised and semi-supervised AU detection}\label{Self-supervised and semi-supervised AU detection}
Recently, there has been a surge in efforts to enhance the accuracy of AU detection through the application of self-supervised learning. In the field of self-supervised learning for AU detection, extracting supervised knowledge from extensive unlabeled input predominantly necessitates auxiliary tasks~\citep{yan:et.al:2021:ACM}.~\citet{wiles:et.al:2018:BMVC-a} introduced the Fab-Net, leveraging facial movement transformations between consecutive frames as the supervisory signal to learn facial embedding through reconstruction loss. Inspired by Fab-Net,~\citet{li:et.al:2019:CVPR} modified facial actions and head poses to decouple facial features from head posture. However, most of these endeavors focus solely on global facial features, overlooking some AU-related knowledge, such as the distinctive properties of AUs, including locality and relevance. Learning only global features has limited efficacy in AU detection since each AU correlates with specific facial muscles and does not manifest independently.

Additionally, several self-supervised methods have been developed to generate potent visual representations for single-object detection through contrastive learning.~\citet{chen:et.al:2020:ICML} proposed SimCLR, a framework training a base encoder network and a projection to maximize agreement using a contrastive loss.~\citet{he:et.al:2020:CVPR} introduced MoCo, incorporating a momentum network to store a queue of negative samples for efficient contrastive learning.~\citet{lu:et.al:2020:BMVC} utilized temporal consistency for feature representation learning through contrastive learning.~\citet{chen:et.al:2021:CVPR} achieved general representations without negative sample pairs, avoiding undesired collapsing solutions via a stop-gradient operation.~\citet{li:et.al:2023:TIP} formulated a contrastive self-supervised learning method for encoding discriminative AU representations from an abundant supply of unlabeled facial videos. However, the prevalent use of self-supervised tasks based on random crops or temporal consistency in contrastive approaches can lead to insufficient AU representations, underutilizing the benefits of static datasets.

Furthermore, the acquisition of reliable AU labels is challenging and highly-cost, and often tainted by noise and errors. Consequently, many semi-supervised frameworks seek supervisory information from alternative sources rather than relying solely on AU labels.~\citet{zhang:et.al:2019:CVPR} proposed a context-aware attention mechanism for joint AU intensity estimation, but its application is limited to a specific dataset.~\citet{cui:et.al:2020} introduced a constraint optimization method encoding prior knowledge on expression-AU probabilistic dependencies into a Bayesian Network, promoting the generalization ability across different datasets.~\citet{chang:et.al:2022:CVPR} utilized AU labeling rules defined by the Facial Action Coding System to devise a knowledge-driven self-supervised representation learning framework for AU detection.

In contrast to the aforementioned methods, our framework introduces an AU domain separation and reconstruction framework, facilitating the transfer of AU knowledge from the source domain to the target domain. This approach offers adaptability to various domains related to the task and can be widely implemented on static datasets.

\subsection{AU domain separation and reconstruction} \label{AU domain separation and reconstruction}
The AU domain separation and construction represents a prevalent method for transferring knowledge between source and target domains. A common strategy involves employing an adversarial loss with a domain discriminator to render the features of both domains indistinguishable.~\citet{ganin:et.al:2016} introduced the Domain-Adversarial Neural Network, a shared network facilitating the learning of domain-invariant features. In contrast,~\citet{tzeng:et.al:2017:CVPR} proposed the Adversarial Discriminative Domain Adaptation method, pre-training a network on the source domain and refining it on the target domain to minimize adversarial loss between fixed source-domain features and trainable target-domain features. Despite the effectiveness of these methods in domain separation and construction, enforcing feature domain invariance proves impractical for AU detection due to the potential removal of AU-related information intertwined with head poses, causing domain gaps.

An alternative involves translating source images into target-style images.~\citet{zheng:et.al:2018:ECCV} presented a method translating rendered images into the real image domain, incorporating identity mapping regularization for real input images.~\citet{lee:et.al:2018:ECCV} separated representations for translating images by embedding them into a specialized attribute space for a specific domain and a content space that remains invariant across domains, capturing shared information.~\citet{wang:et.al:2018:ACM} utilized a generative adversarial network~\citep{goodfellow:et.al:2014:generative} to synthesize images resembling the target image while retaining AU patterns from the source image. This approach represents a pioneering effort in AU detection via AU domain separation and construction. However, limitations arise as the processed source and target images exhibit similar expressions and are both constrained, focusing solely on the image differences of AU patterns. 

Additionally,~\citet{shao:et.al:2022:TAC} proposed an end-to-end unconstrained facial AU detection framework based on domain adaptation, transferring accurate AU labels from a constrained source domain to an unconstrained target domain by leveraging AU-related facial landmarks. However, obtaining a substantial amount of accurate facial landmark ground truths during the training process, particularly in the target domain, proves highly impractical.

Compared to the previous AU domain separation and construction methods, our method implements the domain adaptation of AU features through mixing AU-related features from source domain with background features from target domain. This first merit of our method is that it can extract the key features accurate to each pixel of the internal face, which preserves the data distribution of facial features from source domain as much as possible, and omits the step of removing head poses to avoid the problem of losing AU-related information in the methods based on learning domain-invariant features~\citep{ganin:et.al:2016,tzeng:et.al:2017:CVPR}. Moreover, our method enables to map the source domain images with AU labels into an unconstrained domain preserving the appearances of target domain images such as pose, illumination, occlusion, and background, which avoids the limitation of only transferring images with similar domains in the methods based on translating the image style~\citep{zheng:et.al:2018:ECCV,lee:et.al:2018:ECCV,wang:et.al:2018:ACM}. At the same time, the accurate facial landmark coordinates in target domain needed in the training process can be generated by our introduced facial landmark detection branch from target domain compared to ADLD~\citep{shao:et.al:2022:TAC}, effectively solving the problem that it is difficult to obtain a large amount of accurate facial landmark ground truths.

\subsection{Multi-task learning frameworks for AU detection}\label{Multi-task learning frameworks for AU detection}
Due to the correlation between different facial tasks, deep multi-task learning has been widely used in this field. Firstly, facial landmarks are closely related to AUs, as the former can be used to predefine the positions of AUs. Therefore, there have been many AU detection frameworks based on facial landmark detection. The current facial landmark detection methods are mainly based on two branches, that are coordinate regression-based~\citep{kazemi:et.al:2014:CVPR,zhang:et.al:2015:TPAMI,he:et.al:2017:FG,wu:et.al:2018:CVPR,shao:et.al:2020:Neurocomputing,wan:et.al:2021:TIP,wan:et.al:2022:TNNLS} and heatmap regression-based~\citep{yang:et.al:2017:CVPR,dong:et.al:2018:CVPR,liu:et.al:2019:CVPR,wang:et.al:2020:TPAMI,wan:et.al:2023:TIP}. The heatmap regression-based methods are more accurate at present, which can output larger feature maps and have strong spatial generalization ability. However, the coordinate regression-based methods are end-to-end total differential and have faster training and reasoning speed, which are also more suitable for multi-task learning frameworks.~\citet{shao:et.al:2021:IJCV} was the first in amalgamating AU detection and facial landmark detection, leveraging the inherent similarities between these tasks to bolster performance in each.~\citet{wang:et.al:2023:TAC} introduced an innovative multi-task dual learning framework aimed at concurrently executing facial landmark detection and AU recognition, capturing the relationships between facial landmarks and AUs at both the feature and label levels.

In addition, certain AU detection frameworks utilize facial landmarks to localize crucial facial regions and extract local features relevant to AUs.~\citet{li:et.al:2017:CVPR} proposed a deep learning framework for AU detection incorporating region of interests (ROIs) adaptation, integrated multi-label learning, and optimal LSTM-based temporal fusing.~\citet{li:et.al:2017:CVPR} improved and cropped ROIs with roughly extracted facial landmark information for AU detection.~\citet{yang:et.al:2023:ICONIP} combined basic facial features, attention-based landmark features, and contrastive learning to enhance AU detection performance.

Furthermore, several frameworks are often integrated with downstream tasks of AU detection, such as facial expression recognition, facial attribute detection and stuff like that.~\citet{tallec:et.al:2022:TAC} introduced a multi-task learning approach incorporating task order optimization and prediction for simultaneous detection of facial attributes and AUs. ~\citet{zhou:et.al:2022:IJMLC} developed a multi-task-based facial analysis system integrating various facial attributes, including facial landmarks, head pose, gender, and emotion.~\citet{zhou:et.al:2022:IJMLC} proposed a meta auxiliary learning method that automatically selects highly related facial expression samples by learning adaptive weights for the training facial expression samples in a meta-learning fashion.

Motivated by these methodologies, we have undertaken a novel approach by combining AU domain separation and reconstruction with facial landmark detection for AU detection. Notably, this integration of tasks represents an innovative direction that has not been explored previously.

\FloatBarrier
\section{Our Proposed Method} \label{Our Proposed Method}

\subsection{Overview}\label{Overview}
\begin{figure*}[tbp]
	\centering
	\includegraphics[width=0.92\textwidth,height=0.77\textheight,keepaspectratio]{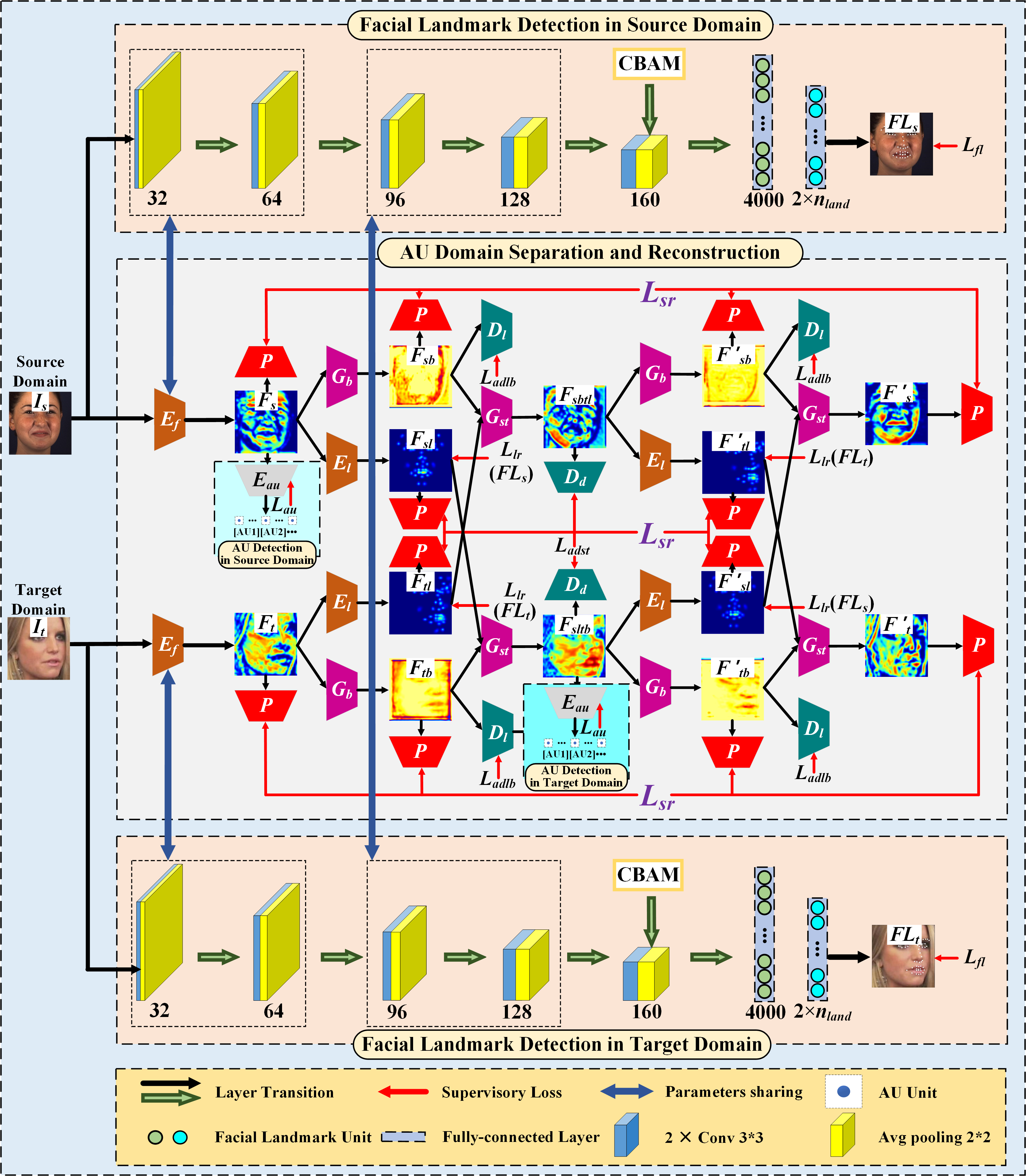}
	\caption{The overall architecture of our proposed multi-task learning model. For AU detection in source domain and target domain, we jointly learn two tasks named AU separation and reconstruction and facial landmark detection to solve the problem of insufficient labels in the wild and weak generalization performance in laboratory scenes. The same features and modules that have the same structure are marked by the same color, and parameters of $G_{b}$, $E_{l}$, $G_{st}$, $D_{l}$, $E_{au}$ and $P$ are shared by source-domain and target-domain input images.}
	\label{fig:2.png}
\end{figure*}

Fig.~\ref{fig:2.png} shows the overall architecture of our proposed multi-task learning model, which is consisted of three branches: two facial landmark detection branches, and an AU domain separation and reconstruction branch. 

Initially, we begin the training of both facial landmark detection branch in source domain and in target domain with two images $I_{s}$ and $I_{t}$ to get 49 facial landmark coordinates in both source domain and target domain and pretraining models for two facial feature extractors $E_{f}$($I_{s}$, $F_{s}$) and $E_{f}$($I_{t}$, $F_{t}$). After these two branches have been pre-trained, the source image $I_{s}$ and the target image $I_{t}$ are put into the domain separation and reconstruction branch to get AU features $F_{s}$ and $F_{t}$. $E_{f}$($I_{s}$, $F_{s}$) and $E_{f}$($I_{t}$, $F_{t}$) will be loaded by initialized parameters from the modules of two facial landmark detection branches. After that, three branches are trained simultaneously. 

The AU domain separation and reconstruction branch is consisted of two branches, a source domain branch and a target domain branch. Each branch is separated and reconstructed twice. After putting two images $I_{s}$ and $I_{t}$ into $E_{f}$($I_{s}$, $F_{s}$) and $E_{f}$($I_{t}$, $F_{t}$), we get two initial facial features $F_{s}$ and $F_{t}$. Then, facial texture encoders $E_{l}$($F_{s}$, $F_{sl}$), $E_{l}$($F_{t}$, $F_{tl}$) and a background feature generator $G_{b}$($F_{s}$, $F_{sb}$), $G_{b}$($F_{t}$, $F_{tb}$) are used to seperate $F_{s}$ and $F_{t}$ into landmark-related features $F_{sl}$ and $F_{tl}$, and background features $F_{sb}$ and $F_{tb}$. As shown in Fig.~\ref{fig:2.png}, the acquisitions of landmark-related features $F_{sl}$ and $F_{tl}$ are supervised by facial landmark labels $FL_{s}$ and $FL_{t}$ from the training of facial landmark detection branches in source domain and target domain, which make our framework become an end-to-end trainable one. A minimax game theory is also employed in the process of feature separation while generators $G_{b}$($F_{s}$, $F_{sb}$) and $G_{b}$($F_{t}$, $F_{tb}$) attempt to generate background features $F_{sb}$ and $F_{tb}$, while discriminators $D_{l}$($F_{sb}$) and $D_{l}$($F_{tb}$) aim to detect landmark-related features in $F_{sb}$ and $F_{tb}$. After that, generators $G_{st}$($F_{sb}$, $F_{sbtl}$) and $G_{st}$($F_{tb}$, $F_{sltb}$) are applied to generate new reconstructed features $F_{sltb}$ and $F_{sbtl}$ by mixing $F_{sl}$ and $F_{tb}$, $F_{sb}$ and $F_{tl}$ together. Discriminators $D_{d}$($F_{sbtl}$) and $D_{d}$($F_{sltb}$) are used to recognize generated features from other domains. Finally, another round of domain separation and reconstruction which is the same as the first round is introduced to $F_{sltb}$ and $F_{sbtl}$ to obtain cross-cyclically reconstructed original features $F_{s}^{\prime}$ and $F_{t}^{\prime}$. 

Apart from that, we propose a pixel-level feature alignment scheme. To better realize the reconstruction process, we introduce projectors $P$($\cdot$) with simple structure and an improved self-reconstruction loss ${L}_{sr}$. Projectors $P$($\cdot$) first amplify the receptive field of feature pixel elements by increasing the depth of neural networks, then ${L}_{sr}$ aligns similar features from separated and reconstructed process pixel by pixel. Different from the original self-reconstruction loss~\citep{shao:et.al:2022:TAC,bhattacharya:et.al:2020:ECCV} with only two pairs of restored features ${F}_{s}^{\prime}$, ${F}_{s}$ and ${F}_{t}^{\prime}$ and ${F}_{t}$, ${L}_{sr}$ is now working on six pairs of features after adding four pairs of intermediate supervisors ${F}_{sl}^{\prime}$ and ${F}_{sl}$, ${F}_{sb}^{\prime}$ and ${F}_{sb}$, ${F}_{tl}^{\prime}$ and ${F}_{tl}$, ${F}_{tb}^{\prime}$ and ${F}_{tb}$. This makes reconstructed features much closer to original ones, increasing the accuracy of AU detection.

The AU feature extractors ${E}_{au}$($F_{s}$) and ${E}_{au}$($F_{sltb}$) are trained for AU detection from source domain and target domain, which essentially rely on a semi-supervised learning framework supervising images from both source domain and target domain by using only AU labels from source domain. This domain adaptive method is also useful to other facial tasks such as facial expression recognition or analysis. More details will be introduced in the following. 

\subsection{Multi-task learning strategy}
As shown in Fig.~\ref{fig:2.png}, our proposed AU detection framework adopts the multi-task learning strategy, jointly learning with both AU domain separation and reconstruction task and facial landmark detection. We introduce two facial landmark detection branches from both source domain and target domain to solve the problem of insufficient facial landmark labels when obtaining landmark-related features in AU domain separation and reconstruction branch. This dual domains facial landmark detection framework can also replace ground truths from other facial toolkits with our training facial landmark lables, which avoids the situation that it is unable to get high-precision landmarks in real scenario. Another advantage of this approach is that it enables weight sharing of partial modules among three facial tasks with similar facial features, which can mutually guide each other during training. 

The training strategy consists of two stages. The first stage is training two facial landmark detection branches. Each branch comprises five integral modules and two fully-connected layers. For each module, there are two 3$\times$3 convolutional layers with a stride of 1 and padding of 1, followed by a 2$\times$2 average-pooling layer. The channels of convolutional layers in five modules are respectively 32, 64, 96, 128 and 160. After that, the CBAM~\citep{Woo:et.al:2018:ECCV} module is appended to the last integral module. At the end of the branch, two fully-connected layers predict facial landmarks trained by ${L}_{fl}$ as
\begin{equation} 
	{L}_{fl}(y,\hat{y})=-\frac{1}{2s^{2}}\sum_{i=1}^{n_{land}}[(y_{2i-1}-\hat{y}_{2i-1})^{2}+(y_{2i}-\hat{y}_{2i})^{2}],
\end{equation}
where $y_{2i-1}$ and $y_{2i}$ represent the actual x-coordinate and y-coordinate of the i-th facial landmark, while $\hat{y}_{2i-1}$ and $\hat{y}_{2i}$ represent the corresponding predicted values. The term ${n}_{land}$ denotes the total number of facial landmarks, and s represents the inter-ocular distance of two eyes used for normalization. 

After the training of two facial landmark detection branches, we separately put both the source image $I_{s}$ and the target image $I_{t}$ into two feature extraction modules $E_{f}$($I_{s}$, $F_{s}$) and $E_{f}$($I_{t}$, $F_{t}$) with initialized parameters provided by the first two integral modules of two facial landmark detection branches, since $E_{f}$($I_{s}$, $F_{s}$) and $E_{f}$($I_{t}$, $F_{t}$) have the same structures as the two integral modules. Then we train three branches, including two facial landmark detection branches and an AU domain separation and reconstruction branch simultaneously. In addition, the third and fourth integral modules of two facial landmark detection branches also share parameters during the whole training process. The advantage of this way is that the number of weight updates that must be carried out in the process of backpropagation is reduced, and the generalization performance of modules from different domains can be improved by mutually guiding each other during training. Furthermore, the initialized pretraining parameters for facial feature extraction also help $E_{f}$($I_{s}$, $F_{s}$) and $E_{f}$($I_{t}$, $F_{t}$) warm up before training, which are able to promote the feature separation process. 

\subsection{AU Domain Feature Separation and Reconstruction}
The AU domain feature separation and reconstruction process contains two branches: the source domain branch and the target domain branch. Each branch is separated and reconstructed twice. Therefore, for the convenience of expression, we only take the target domain branch as an example to explain the whole AU domain feature separation and reconstruction process. 

To begin with, our goal is to isolate the initial facial feature $F_{t}$ into two features, the landmark-related feature $F_{tl}$ and the background feature $F_{tb}$. The former one includes the inner-landmark information and the latter one excludes the facial inner-landmark information. Due to the fact that adversarial learning based on cross entropy loss ~\citep{goodfellow:et.al:2014:generative} is widely used for feature separation, we first convert the facial landmark detection task from a regression task~\citep{zhang:et.al:2015:TPAMI,shao:et.al:2020:Neurocomputing} into a classification task~\citep{honari:et.al:2016:CVPR,xiao:et.al:2016:ECCV} to get $F_{tb}$. To be more specific, we transform the square feature map of each landmark with a size of $d$$\times$$d$$\times$1 into independent an one-hot coding sequence with a size of $d^{2}$$\times$1, since each position on the square feature map is regarded as a binary classification problem. In this multi-label binary classification problem, only the sample of the facial landmark position is a positive one. $d$ represents the side length of the square feature map and $d^{2}$ is the length of the one-hot coding sequence. We assume that $p^{t}_{2i}$ and $p^{t}_{2i-1}$ are the coordinates of the i-th facial landmark and ${y}^{t}_{i}$ is the index of the facial landmark position in the one-hot coding sequence after transformation. The transformation is defined as
\begin{equation}
	{y}^{t}_{i}=(\lfloor p^{t}_{2i}\cdot d/l\rfloor-1)\cdot d+\lfloor p^{t}_{2i-1}\cdot d/l\rfloor,
\end{equation}
where the notation $\lfloor \cdot \rfloor$ denotes the operation of taking a largest integer not greater than the number. $l$ represents the crop size of the input image $I_{t}$. The ground truth is assigned a value of 1 at the facial landmark position and 0 at all other positions.

By regarding the facial landmark detection as a multi-label binary classification problem, we utilize facial texture encoder $E_{l}$ for extracting landmark-related feature:
\begin{equation} \label{lr}
	\begin{aligned}
		&{L}_{lr}({E}_{l}, {F}_{t}, {F}_{tl})=  \\ 
		&\mathbb{E}_{{f}^{t}\sim{F}_{t}}[\frac{1}{n}\sum^{n}_{i=1}\sum^{d^{2}}_{k=1}\mathbb{1}_{[k=y^{t}_{i}]}log(\sigma(E^{(i,k)}_{l}(f^{t})))], 
	\end{aligned}
\end{equation}
where ${f}^{t}$ is the random variable subject to the distribution of ${F}_{t}$. $E^{(i,k)}_{l}(\cdot)$ represents the k-th value in the square feature map of the i-th landmark, $\mathbb{1}_{[\cdot]}$ is the indicator function, and $\sigma(\cdot)$ represents the softmax function. The significance of Eq.~\eqref{lr} is to encourage $E_{l}$($F_{t}$, $F_{tl}$) to have the highest value $\sigma(E^{(i,k)}_{l}(f^{t}))$ at the position ($\lfloor p^{t}_{2i-1} \cdot d/l\rfloor,\lfloor p^{t}_{2i} \cdot d/l\rfloor$) while having no value at other positions. To make the landmark-related feature $F_{tl}$ include facial landmark information, we can formalize $F_{tl}$ as $\oplus^{n}_{i=1}\sigma(E^{(i)}_{l}(f^{t}))$, where $\oplus$ represents the per-pixel addition.

To generate background features without facial landmark information, we introduce the multi-label minimax game theory into the feature separation process, which is also a kind of adversarial learning. Specifically, we train $D_{l}$($F_{tb}$) by minimizing
\begin{equation} 
	\begin{aligned}
		\mathbb{E}_{{f}^{t}\sim{F}_{t}}[\frac{1}{nd^{2}}\sum^{n}_{i=1}\sum^{d^{2}}_{k=1}(\mathbb{1}_{[{k}\neq{y}^{t}_{i}]}\parallel{D}^{(i,k)}_{l}(G_{b}({f}^{t}))\parallel^{2}_{2}+\\
		\mathbb{1}_{[{k}={y}^{t}_{i}]}\parallel{D}^{(i,k)}_{l}(G_{b}({f}^{t}))-1\parallel^{2}_{2})],\\
	\end{aligned}
\end{equation}
where the function of ${D}^{(i,k)}_{l}(\cdot)$ is similar to ${E}^{(i,k)}_{l}(\cdot)$. We encourage ${D}_{l}$($F_{tb}$) to distinguish 1 at the facial landmark position, while it distinguishes 0 at all other positions. On the contrary, we train $G_{b}$ by minimizing
\begin{equation} 
	\begin{aligned}
		\mathbb{E}_{{f}^{t}\sim{F}_{t}}[\frac{1}{nd^{2}}\sum^{n}_{i=1}\sum^{d^{2}}_{k=1}\parallel{D}^{(i,k)}_{l}(G_{b}({f}^{t}))-\frac{1}{{{d}^{2}}}\parallel^{2}_{2}],
	\end{aligned}
\end{equation}
where ${G}_{b}$ try its utmost to generate background feature $F_{tb}$, while $D_{l}$($F_{tb}$) is designed to assign a uniform detection probability of $1/{{d}^{2}}$ to all potential facial landmark locations. The multi-label minimax game theory of adversarial learning plays an important role for generating background feature $F_{tb}$. The facial landmark adversarial loss $L_{adlb}$ is defined as
\begin{equation} 
	\begin{aligned}
		&{L}_{adlb}({G}_{b}, {D}_{l}, {F}_{t}, {F}_{tb})= \\
		&\mathbb{E}_{{f}^{t}\sim{F}_{t}}[\frac{1}{n}\sum^{n}_{i=1}\sum^{d^{2}}_{k=1}\mathbb{1}_{[k=y^{t}_{i}]}log(\sigma({D}^{(i,k)}_{l}(G_{b}(f^{t}))))],
	\end{aligned}
\end{equation}
where $G_{b}(f^{t})$=${F}_{tb}$. $L_{adlb}$ is a kind of multi-class cross entropy loss.

We want to get generating feature ${F}_{sltb}$ that both have AU labels and a background from target domain. To encourage ${F}_{sltb}$ to be indistinguishable from other features whose backgrounds are from target domain, we introduce another adversarial loss ${L}_{adst}$ for the generation of ${F}_{sltb}$:
\begin{equation} 
	\begin{aligned}
		&{L}_{adst}({E}_{l}, {G}_{b}, {G}_{st}, {D}_{d}, {F}_{t}, {F}_{sltb})= \\
		&\mathbb{E}_{{f}^{t}\sim{F}_{t}}[log{D}_{d}({f}^{t})]+\mathbb{E}_{{f}^{sltb}\sim{F}_{sltb}}[log(1-{D}_{d}(f^{sltb}))].
	\end{aligned}
\end{equation}
To achieve stable feature reconstruction, we use the least-squares loss for ${L}_{adst}$. Specially, we train ${D}_{d}$ by minimizing $\mathbb{E}_{{f}^{t}\sim{F}_{t}}[\parallel{D}_{d}({f}^{t})-1\parallel^{2}_{2}]+\mathbb{E}_{{f}^{sltb}\sim{F}_{sltb}}[\parallel{D}_{d}(f^{sltb})\parallel^{2}_{2}]$, and train ${G}_{st}$ by minimizing $\mathbb{E}_{{f}^{sltb}\sim{F}_{sltb}}[\parallel{D}_{d}(f^{sltb})-1\parallel^{2}_{2}]$. The generator $G_{st}$ is applied to generate new reconstructed features $F_{sltb}$ from target domain and the discriminator $D_{d}$ is used to recognize generated features which are not from target domain. 

\subsection{Pixel-level Feature Alignment scheme}
To maintain the consistency of features obtained from two separation and reconstruction processes, we propose a pixel-level feature alignment scheme, which is realized by 12 projectors $P$(${F}_{s}$), $P$(${F}_{t}$), $P$(${F}_{sl}$), $P$(${F}_{tl}$), $P$(${F}_{sb}$), $P$(${F}_{tb}$), $P$(${F}_{s}^{\prime}$), $P$(${F}_{t}^{\prime}$), $P$(${F}_{sl}^{\prime}$), $P$(${F}_{tl}^{\prime}$), $P$(${F}_{sb}^{\prime}$), $P$(${F}_{tb}^{\prime}$) with simple structure and an improved self-reconstruction loss ${L}_{sr}$. Specifically, four pairs of intermediate supervisors ${F}_{sl}^{\prime}$ and ${F}_{sl}$, ${F}_{sb}^{\prime}$ and ${F}_{sb}$, ${F}_{tl}^{\prime}$ and ${F}_{tl}$, ${F}_{tb}^{\prime}$ and ${F}_{tb}$ are added to ${L}_{sr}$ comparing to the original self-reconstruction loss~\citep{shao:et.al:2022:TAC,bhattacharya:et.al:2020:ECCV} with only two pairs of restored features ${F}_{s}^{\prime}$, ${F}_{s}$ and ${F}_{t}^{\prime}$ and ${F}_{t}$. The self-reconstruction loss mainly refer to the thought of contrastive learning loss, which utilize ${L}_{1}$ loss with strong robustness: 

\begin{equation}
	\begin{aligned}
	&{L}_{c}= \\
	&\frac{1}{{h}_{s} \times {w}_{s}}\sum_{{h}_{s},{w}_{s}}\parallel{F}_{s}^{\prime}-{F}_{s}\parallel +\frac{1}{{h}_{sl} \times {w}_{sl}}\sum_{{h}_{sl},{w}_{sl}}\parallel{F}_{sl}^{\prime}-{F}_{sl}\parallel+\\
	&\frac{1}{{h}_{sb} \times {w}_{sb}}\sum_{{h}_{sb},{w}_{sb}}\parallel{F}_{sb}^{\prime}-{F}_{sb}\parallel
	+\frac{1}{{h}_{t} \times {w}_{t} }\sum_{{h}_{t},{w}_{t}}\parallel{F}_{t}^{\prime}-{F}_{t}\parallel+ \\
	&\frac{1}{{h}_{tl} \times {w}_{tl} }\sum_{{h}_{tl},{w}_{tl}}\parallel{F}_{tl}^{\prime}-{F}_{tl}\parallel +\frac{1}{{h}_{tb} \times {w}_{tb} }\sum_{{h}_{tb},{w}_{tb}}\parallel{F}_{tb}^{\prime}-{F}_{tb}\parallel,
	\end{aligned}
\end{equation}
where ${h}$ and ${w}$ denote the spatial locations of each features. Although ${L}_{1}$ loss is simple in form, it is very sensitive to large errors between the predicted value and the actual value. Each projector $P$($\cdot$) is used to amplify the receptive field of feature pixel elements by a relatively simple structure, which is consisted of a 3$\times$3 convolutional layer between two 1$\times$1 convolutional layer with a stride of 1 and padding of 1. These projectors are also efficient enough to ensure accurate alignment and match the dimensions between each pairs of similar facial features. In fact, this simple loss has already been utilized before~\citep{chen:et.al:2022:knowledge,adriana:et.al:2015:fitnets,yang:et.al:2020:knowledge}, but we are actually tring to reveal the potential value of the additional intermediate supervisors rather than designing a complicated loss for feature alignment. The advantage of this approach is that it achieves pixel-by-pixel alignment of the face-background features and the four additional intermediate supervisions greatly improve the reconstruction process.

\subsection{Facial AU detection}
The AU detection loss is employed to detect ${F}_{s}$, which represents the initial AU features from source domain, as well as ${F}_{sltb}$, which refers to the reconstructed AU feature mixing landmark-related feature ${F}_{sl}$ from source domain and background feature ${F}_{tb}$ from target domain together. Traditional supervised AU detection methods use the weighted multi-label cross entropy loss~\citep{shao:et.al:2021:IJCV,shang:et.al:2023:PRL} for AU detection. However, we find that AU labels in BP4D and DISFA are very imbalanced. For most AUs, the number of negative samples is much larger than the number of positive ones. To solve this problem of multi-label binary classification with imbalanced samples, we propose a weighted asymmetric loss ${L}_{au}$ to calculate the values between ground-truth occurrence values and predicted occurrence probabilities generated by ${E}_{au}$($F_{s}$) and ${E}_{au}$($F_{sltb}$), which is formulated as
\begin{equation} 
	\begin{aligned}
	&{L}_{au}(y,\hat{y}) = \\
	&-\frac{1}{n_{au}}\sum_{i=1}^{n_{au}}w_{i}[y_{i}log\hat{y}_{i}+(1-{y}_{i}){(p)}^{\beta}log(1-\hat{y}_{i})],
	\end{aligned}
\end{equation}
where $y_{i}$ represents the ground-truth occurrence value (0 or 1) of the $i$-th AU. The corresponding predicted occurrence probability is denoted as $\hat{y}_{i}$. The weight $w_{i}$ is calculated as $w_{i}=\frac{(1/r_{i})n_{au}}{\sum_{i=1}^{n_{au}}(1/r_{i})}$, which is used to ensure balanced training. It permits reducing the impact of loss values associated with those AUs that are more commonly activated in the training set. Consequently, loss values arising from less frequently inactivating AUs are given greater significance during the training process. $r_{i}$ represents the occurrence frequency of the $i$-th AU, and $n_{au}$ is the total number of AUs.
	
Inspired by the asymmetric loss~\citep{ridnik:et.al:2021:ICCV,luo:et.al:2022:IJCAI}, we add the expression ${(p)}^{\beta}$ at the center of (1-${y}_{i}$)${(p)}^{\beta}$log(1-$\hat{y}_{i}$). $p$ represents the shifted probability, which is formulated as:
\begin{equation} \label{p} 
	\begin{aligned}
		p=max(\hat{y}_{i}-\gamma, 0).
	\end{aligned}
\end{equation} 

An additional asymmetric mechanism, referred to as probability shifting, is introduced to effectively eliminate easily identifiable negative samples when their associated probabilities reach exceedingly low levels. The probability margin $\gamma \geq 0$ is a tunable hyperparameter. Furthermore, to underscore the significance of positive samples, a parameter $\beta>0$ is typically established to diminish the influence of easily classified negatives (having low probability, $\hat{y}_{i}\ll$0.5) within the loss. This adjustment directs the training process to prioritize more challenging samples, thereby enhancing the model's ability to discern meaningful features from positive samples, despite their inherent rarity. This approach ensures superior control over the contribution of both positive and negative samples to the loss, thereby fostering a more effective learning process. For the negative samples(${y}_{i}=0$) of each AU, the asymmetric loss is:
 
\begin{equation} 
	\begin{aligned}
		\mathcal{L}=(p)^{\beta}log(1-p).
	\end{aligned}
\end{equation} 

In Eq.~\eqref{p}, the probability shifting is analogous to a rightward translation of the loss by a factor denoted as $\gamma$, leading to the condition $\mathcal{L}=0$ when $p < \gamma$. $\mathcal{L}$ facilitates the incorporation of two distinct forms of asymmetry mechanisms, aimed to change the magnitude of $\mathcal{L}$ with respect to positive and negative samples. This is achieved through the modulation of the loss via two key parameters: the focusing parameter $\beta$ and the probability margin $\gamma$.

As the updating of network weights is contingent upon the gradient of the loss concerning the input logit $\theta$, our subsequent endeavor involves an analysis of the loss gradients for negative samples in $\mathcal{L}$:
\begin{equation} 
	\begin{aligned}
		&\frac{\partial\mathcal{L}}{\partial\theta}=\frac{\partial\mathcal{L}}{\partial\hat{y}_{i}}\frac{\partial\hat{y}_{i}}{\partial\theta} \\
		&=(p)^{\beta}[\frac{1}{1-p}-\frac{\beta log(1-p)}{p}]\hat{y}_{i}(1-\hat{y}_{i}),
	\end{aligned}
\end{equation} 

Fig.~\ref{fig:3.png} illustrates the normalized gradients of $\mathcal{L}$ across various values of $\beta$ and $\gamma$. According to Fig.~\ref{fig:3.png}, the negative samples can be broadly categorized into three distinct loss regimes:

\begin{figure}[tbp]
	\centering
	\includegraphics[width=0.99\linewidth]{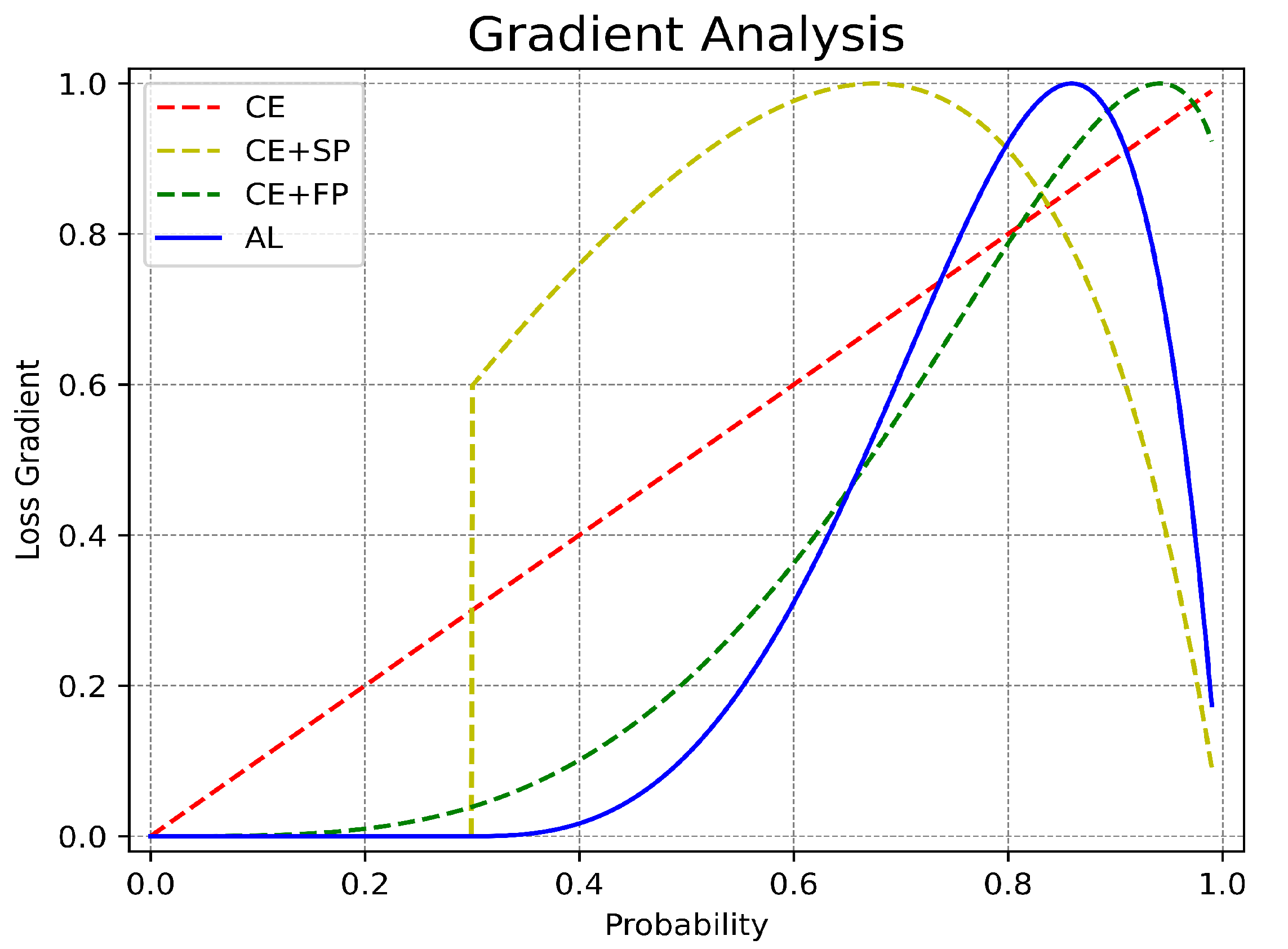}
	\caption{Gradient Analysis of $\mathcal{L}$, where $\hat{y}_{i}$ is Probability and $\frac{\partial\mathcal{L}}{\partial\theta}$ is Loss Gradient. CE represents Cross-Entropy ($\gamma=0$, $\beta=0$). CE+SP represents Cross-Entropy with shifted probability ($\gamma>0$, $\beta=0$). CE+FP represents Cross-Entropy with focusing parameter ($\gamma=0$, $\beta>0$). AL represents asymmetrics loss ($\gamma>0$, $\beta>0$).}
	\label{fig:3.png}
\end{figure}

1.Very easy negatives, with $\hat{y}_{i}$ < $\gamma$, that should be ignored, to prioritize the emphasis on harder samples.

2.Typical negative samples, with $\hat{y}_{i} \geq \gamma$, warrant non-linear attenuation when their probabilities are low.

3.Very hard negative samples, with $\hat{y}_{i}$ > $\hat{y}_{i}^{*}$, where $\hat{y}_{i}^{*}$ is defined as the point where $\frac{d}{d \hat{y}_{i}}(\frac{d \mathcal{L}}{d \theta})=0$, are indicative of potential mislabeling. Given the propensity of neural networks for memory effects, even a small mislabeling rate among negative samples can significantly impact training, especially in scenarios with highly imbalanced datasets. Consequently, the careful rejection of potentially mislabeled samples prove advantageous.

As depicted in Fig.~\ref{fig:3.png}, it becomes evident that the confluence of the two asymmetry mechanisms, focusing and shifted probability, bestows a comprehensive set of advantages crucial for handling imbalanced datasets: ignoration of very easy nagetive samples, non-linear attenuation of typical negative samples, rejection of mislabeled nagetive samples and continuous loss gradients.

Combining all the losses, we yield the overall loss
\begin{equation} \label{overall}
	\begin{aligned}
		&L_{all}={\lambda}_{lr}{L}_{lr}+{\lambda}_{adlb}{L}_{adlb} \\ 
		&+{\lambda}_{adst}{L}_{adst}+{\lambda}_{au}{L}_{au}+{\lambda}_{fl}{L}_{fl}+{\lambda}_{sr}{L}_{sr},
	\end{aligned}
\end{equation}
where the hyperparameters ${\lambda}_{(.)}$ are utilized to control the relative significance of each loss term in our proposed multi-task learning strategy. Moreover, Eq.~\eqref{overall} is used for both the source domain and the target domain, and the parameters of modules with the same structure in each domain are shared. It is important to note that our framework is designed to be trainable in an end-to-end manner. This implies that all network modules are simultaneously trained in a joint optimization process, ensuring a comprehensive and integrated learning approach.

\FloatBarrier
\section{Experiments} \label{Sec:Experiments}
\subsection{Experiment Settings} \label{Experiment Settings}
\subsubsection{Datasets}In our experiments, we totally utilize three widely-used datasets for AU detection. For the supervised source domain, we utilize BP4D~\citep{zhang:et.al:2014:IVC} and DISFA~\citep{mavadati:et.al:2013:TAC}. For the unsupervised target domain, we utilize EmotioNet~\citep{fabian:et.al:2016:CVPR}. 

\textbf{BP4D}~\citep{zhang:et.al:2014:IVC} contains 41 participants from 23 females and 18 males. Each person is involved in 8 sessions captured with both 2D and 3D videos. There are about 140,000 frames with AU labels of occurrence or nonoccurence. Each frame is also annotated with 68 landmarks detected by SDM~\citep{xiong:et.al:2013:CVPR}. 

\textbf{DISFA}~\citep{mavadati:et.al:2013:TAC} consists of 27 videos recorded from 15 men and 12 women. Each video has 4,845 frames and the total number of images is 130,788. The frames are annotated with AU intensities from 0 to 5 and 68 landmarks detected by AAM~\citep{cootes:et.al:2001:TPAMI}. The frames with intensities equal or greater than 2 are considered as positive, while others are treated as negative. 

\textbf{EmotioNet}~\citep{fabian:et.al:2016:CVPR} is a dataset containing approximately one million training and validation images that are collected from the Internet. AU labels for the validation images are manually annotated by certified experts. The facial toolkit dlib~\citep{king:et.al:2012:access} is used to annotate all 68 landmarks, and any images without detected landmarks are subsequently excluded from the dataset. We randomly select 200,000 training images as the training set and split the validation images into a validation set (10,544 images) and a test set (10,544 images). 

For all datasets, we employ a three-fold cross-validation methodology for rigorous evaluation.

\subsubsection{Implementation Details} 
Firstly the 68 facial landmarks are transformed into 49 facial internal landmarks, excluding the facial contour landmarks. Then images are augmented through similarity transformation including in-plane rotation, uniform scaling, and translation using 49 facial landmarks to the same resolution of 200$\times$200$\times$3 pixels, and randomly cropped into $l$$\times$$l$$\times$3 pixels with random horizontal flip. In our experimental setup, the crop size $l$ and the side length $d$ of square feature map are set to 176 and 44. 

For the training process, we first simultaneously pretrain two facial landmark detection branches using Adam~\citep{kinga:et.al:2015:ICLR} optimizer. We set $\beta_{1}$ = 0.9, $\beta_{2}$ = 0.999, an initial learning rate of 0.001, a weight decay of 0.0005, a batch size of 8 and 20 epochs. The learning rate decays per 10 epochs, which is up to 0.5 of the original. Secondly, pre-training parameters are respectively loaded to two feature extractors $E_{f}$($I_{s}$, $F_{s}$) and $E_{f}$($I_{t}$, $F_{t}$). We train the whole multi-task learning model using Adam optimizer for up to 12 epochs. For two facial landmark detection branches, we keep the learning rate of 0.0005. For $G_{b}$, $G_{st}$, $D_{l}$ and $D_{d}$, we set $\beta_{1}$ = 0.5, $\beta_{2}$ = 0.9 and a learning rate of 0.00005. For $E_{f}$, $E_{au}$ and $E_{l}$, we set $\beta_{1}$ = 0.95, $\beta_{2}$ = 0.999 and another learning rate of 0.0001. All learning rates remain unchanged during the first 4 epochs and linearly decay during the next 8 epochs. Thirdly, we choose the pre-trained model with the best performance and fine-tune it. To better set appropriate values for the trade-off parameters $\lambda_{lr}$, $\lambda_{adlb}$, $\lambda_{adst}$, $\lambda_{au}$, $\lambda_{fl}$ and $\lambda_{sr}$, we use the control variable method to change only one parameter every time and set $\lambda_{lr}$, $\lambda_{adlb}$, $\lambda_{adst}$, $\lambda_{au}$, $\lambda_{fl}$ and $\lambda_{sr}$ to 0.6, 400, 1.2, 1, 0.1 and 100 respectively by verification experiments. The experimental results of these hyperparameters are shown in Table 7. 

\subsubsection{Evaluation Metrics} BP4D, DISFA, and EmotioNet collectively contain 17 AU labels (AU 1, AU 2, AU 4, AU 5, AU 6, AU 7, AU 9, AU 10, AU 12, AU 14, AU 15, AU 17, AU 20, AU 23, AU 24, AU 25, AU 26). The descriptions of AU centers and their respective location definitions are the same in ADLD~\citep{shao:et.al:2022:TAC}. 

Table.~\ref{tab:1} provides a summary of the occurrence rates of different AUs in the training sets of the source domain. There is variation in the occurrence rates of different AUs, with some having low rates while others have high rates. To tackle the problem of data imbalance, our framework evaluation focuses solely on AUs with occurrence rates exceeding 7$\%$ in the training set of the source domain. Specifically, for the BP4D dataset, we utilize AUs 1, 2, 4, 6, 12, and 17, while for the DISFA dataset, we consider AUs 4, 6, 12, 25, and 26.

\begin{table}[tbp]
	\centering
\small
		\caption{AU occurrence rates ($\%$) of source-domain training datasets. “-” denotes the dataset does not contain this AU. AUs with occurrence rates larger than 7$\%$ are marked in bold.}
		\tabcolsep=0.08cm
		\label{tab:1}
		\begin{adjustbox}{max width=\linewidth}
\begin{tabular}{c c c c c c c c c c c}
			\toprule
			\multicolumn{1}{l}{\multirow{2}{*}{$\textbf{Dataset}$}}&\multicolumn{10}{c}{$\textbf{AU Index}$}\\  [0.1ex] 
			\cline{2-11}
			\multicolumn{1}{l}{}& 1 & 2 & 4 & 5 & 6 & 9 & 12 & 17 & 25 & 26\\ [0.1ex] 
			\hline 
			BP4D & $\textbf{18.4}$ & $\textbf{14.6}$ & $\textbf{19.8}$ & 3.3 & $\textbf{44.0}$ & 5.7 & $\textbf{54.0}$ & $\textbf{34.2}$ & - & - \\ [0.1ex] 
			DISFA & 4.3 & 3.6 & $\textbf{12.2}$ & 0.8 & $\textbf{7.2}$ & 3.1 & $\textbf{12.9}$ & 4.4 & $\textbf{26.2}$ & $\textbf{8.8}$\\ [0.1ex] 
			\bottomrule 
		\end{tabular}
\end{adjustbox}
	
\end{table}

The evaluation metrics of AU detection and facial landmark detection are both shown as follows:

- \textbf{AU detection.} Consistent with state-of-the-art approaches, we employed two widely used frame-based metrics for AU detection: F1-score($\%$) and Accuracy($\%$). F1-score is the harmonic mean of precision and recall, used the most widely in AU detection. For each method, we compute average metrics over all AUs (denoted as Avg.).

- \textbf{Facial landmark detection.} For this task, we use both Mean error ($\%$) and Failure rate ($\%$). The mean error of more than 10\% is considered unqualified.

\subsection{Comparison with State-of-the-Art Methods} \label{Comparison with State-of-the-Art Methods}
We evaluate our approach against current state-of-the-art methods, including AU detection in target domain methods, AU detection in source domain methods and facial landmark detection methods.

\subsubsection{AU Detection in Target Domain}
Due to our framework does not utilize AU labels during the target domain training process, it is essentially a self-supervised learning one. However, to compare with fully-supervised AU detection methods, we introduce AU pseudo labels of the target domain images from facial toolkit OpenFace~\citep{baltrusaitis:et.al:2018:FG} in our framework. Specifically, we add another two AU detection branches after $F_{t}$ and $F_{sltb}$ respectively, supervised by the introduced AU pseudo labels. This also improves the accuracy of AU detection after $F_{s}$ and $F_{sbtl}$, since the parameters of modules with the same structure of the two branches in AU domain separation and reconstruction process are shared.

We compare our AU detection results without pseudo labels in target domain to traditional adversarial domain adaptation methods, including $\textbf{ADDA}$~\citep{tzeng:et.al:2017:CVPR}, $\textbf{DANN}$~\citep{ganin:et.al:2016} which learn domain invariant features, and $\textbf{T$^{2}$Net}$~\citep{zheng:et.al:2018:ECCV}, $\textbf{DRIT}$~\citep{lee:et.al:2018:ECCV},  $\textbf{ADLD}$~\citep{shao:et.al:2022:TAC}, which transfer the source-style images into target-style images. In addition, the results with pseudo labels are compared with $\textbf{ADLD}$ and three fully-supervised AU detection methods $\textbf{J$\hat{\rm \textbf{A}}$A-Net-P}$~\citep{shao:et.al:2021:IJCV}, $\textbf{ARL-P}$~\citep{shao:et.al:2019:TAC}, $\textbf{WSC-P}$~\citep{zhao:et.al:2018:CVPR}. Quantitative results on the target domain dataset EmotioNet when the source domain is BP4D are summarized in Table.~\ref{tab:2}, where our method achieves the best overall performance. 

\begin{table*}[tbp] 
	\centering
\small
		\caption{F1-score of our target domain results and state-of-the-art methods on EmotioNet, where the source domain dataset is BP4D.}
		\label{tab:2}
		\tabcolsep=0.2cm
		\begin{adjustbox}{max width=\linewidth}
\begin{tabular}{c l c c c c c c c}
			\toprule
			\multicolumn{1}{c}{\multirow{2}{*}{$\textbf{Training strategy}$}}&\multicolumn{1}{l}{\multirow{2}{*}{$\textbf{Method}$}}&\multicolumn{6}{c}{$\textbf{AU Index}$}&\multicolumn{1}{c}{\multirow{2}{*}{$\textbf{Avg.}$}}\\ 
			\cline{3-8}
			\multicolumn{1}{l}{}&&$\textbf{1}$&$\textbf{2}$&$\textbf{4}$&$\textbf{6}$&$\textbf{12}$&$\textbf{17}$& \\
			\hline  
			\multicolumn{1}{c}{{\multirow{6}{*}{$\textbf{Self-supervised learning}$}}}&{$\textbf{ADDA}$~\citep{tzeng:et.al:2017:CVPR}}&17.7&5.2&15.3&38.2&58.7&6.2&23.6\\
			\multicolumn{1}{l}{}&{$\textbf{DANN}$~\citep{ganin:et.al:2016}}&16.8&12.7&25.8&28.9&62.5&9.5&26.0\\
			\multicolumn{1}{l}{}&{$\textbf{T$^{2}$Net}$~\citep{zheng:et.al:2018:ECCV}}&9.4&9.6&24.4&45.1&69.5&4.7&27.1\\ 
			\multicolumn{1}{l}{}&{$\textbf{DRIT}$~\citep{lee:et.al:2018:ECCV}}&20.4&7.7&30.9&44.2&67.5&8.3&29.8\\
			\multicolumn{1}{l}{}&{$\textbf{ADLD}$~\citep{shao:et.al:2022:TAC}}&19.8&$\textbf{25.2}$&31.0&58.2&78.3&8.6&36.8\\
			\multicolumn{1}{l}{}&{$\textbf{Ours}$}&$\textbf{24.4}$&24.6&$\textbf{36.8}$&$\textbf{72.0}$&$\textbf{81.2}$&$\textbf{13.4}$&$\textbf{42.1}$\\ 
			\hline 
			\multicolumn{1}{c}{{\multirow{5}{*}{$\textbf{Pseudo-labels supervised learning}$}}}&{$\textbf{WSC-P}$~\citep{zhao:et.al:2018:CVPR}}&22.7&16.5&39.3&55.2&83.4&10.8&38.0\\
			\multicolumn{1}{l}{}&{$\textbf{J$\hat{\rm \textbf{A}}$A-Net-P}$~\citep{shao:et.al:2021:IJCV}}&21.3&19.2&43.6&59.3&$\textbf{85.2}$&9.3&39.7\\
			\multicolumn{1}{l}{}&{$\textbf{ARL-P}$~\citep{shao:et.al:2019:TAC}}&22.3&17.4&48.1&57.5&84.6&11.9&40.3\\ 			
			\multicolumn{1}{l}{}&{$\textbf{ADLD-P}$~\citep{shao:et.al:2022:TAC}}&30.7&26.1&48.1&60.7&77.6&11.5&42.4\\
			\multicolumn{1}{l}{}&{$\textbf{Ours-P}$}&$\textbf{34.2}$&$\textbf{26.9}$&$\textbf{49.7}$&$\textbf{67.5}$&79.7&$\textbf{17.3}$&$\textbf{45.9}$\\ 
			\bottomrule 
		\end{tabular}
\end{adjustbox}
	
\end{table*}

For self-supervised learning frameworks, compared to ADDA and DANN, our method achieves an average increase of 18.5\% and 16.1\%. This occurs because by enforcing domain invariance of features when feeding them into the AU detector $E_{au}$($F_{sltb}$), and essential AU-related information may be overlooked. Compared to T$^{2}$Net and DRIT, our method outperforms 15.0\% and 12.3\%. This is attributed to the ability of our approach to overcome the limitations caused by substantial domain shifts, such as variations in head poses and occlusion distributions, which can negatively impact target-domain AU detection. Furthermore, our method outperforms 5.3\% in F1-score to ADLD, which confirms the effectiveness of multi-task learning strategy, feature alignment scheme and weighted asymmetric loss. 

For pseudo-labels supervised learning frameworks, compared to J$\hat{\rm A}$A-Net-P and ARL-P, our method achieves an average increase of 6.2\% and 5.6\%. This occurs because the complex fully-supervised AU detection modules need to adjust the optimizer hyperparameters and loss weights precisely during training, otherwise AU detection accuracy will be affected. However, the structure of our AU detection module ${E}_{au}$($F_{sltb}$) is relatively simple. Besides, although WSC-P can refine the inaccurate pseudo AU labels, its results are the worst. 

Table.~\ref{tab:3} summarizes the results obtained from EmotioNet when the source domain is DISFA. Our method demonstrates superior performance in overall AU detection compared to other approaches.

\begin{table*}[tbp] 
	\centering
\small
		\caption{F1-score of our target domain results and state-of-the-art methods on EmotioNet, where the source domain dataset is DISFA.}
		\label{tab:3}
		\tabcolsep=0.25cm
		\begin{adjustbox}{max width=\linewidth}
\begin{tabular}{c l c c c c c c}
			\toprule
			\multicolumn{1}{c}{\multirow{2}{*}{$\textbf{Training strategy}$}}&\multicolumn{1}{l}{\multirow{2}{*}{$\textbf{Method}$}}&\multicolumn{5}{c}{$\textbf{AU Index}$}&\multicolumn{1}{c}{\multirow{2}{*}{$\textbf{Avg.}$}}\\ 
			\cline{3-7}
			\multicolumn{1}{l}{}&&$\textbf{4}$&$\textbf{6}$&$\textbf{12}$&$\textbf{25}$&$\textbf{26}$& \\
			\hline  
			\multicolumn{1}{c}{{\multirow{4}{*}{$\textbf{Self-supervised learning}$}}}&{$\textbf{T$^{2}$Net}$~\citep{zheng:et.al:2018:ECCV}}&23.0&37.9&61.6&80.9&28.0&46.3\\ 
			\multicolumn{1}{l}{}&{$\textbf{DRIT}$~\citep{lee:et.al:2018:ECCV}}&38.0&49.0&64.4&76.4&19.2&49.4\\
			\multicolumn{1}{l}{}&{$\textbf{ADLD}$~\citep{shao:et.al:2022:TAC}}&27.0&51.8&$\textbf{73.8}$&88.4&$\textbf{34.2}$&55.0\\
			\multicolumn{1}{l}{}&{$\textbf{Ours}$}&$\textbf{40.0}$&$\textbf{65.4}$&66.0&$\textbf{90.1}$&28.1&$\textbf{57.9}$\\ 
			\hline 
			\multicolumn{1}{c}{{\multirow{4}{*}{$\textbf{Pseudo-labels supervised learning}$}}}&{$\textbf{J$\hat{\rm \textbf{A}}$A-Net-P}$~\citep{shao:et.al:2021:IJCV}}&47.0&59.7&85.6&88.1&34.4&62.9\\
			\multicolumn{1}{l}{}&{$\textbf{ARL-P}$~\citep{shao:et.al:2019:TAC}}&$\textbf{52.5}$&59.6&86.1&89.0&35.5&64.5\\ 			
			\multicolumn{1}{l}{}&{$\textbf{ADLD-P}$~\citep{shao:et.al:2022:TAC}}&49.5&63.6&86.8&88.6&38.3&65.3\\
			\multicolumn{1}{l}{}&{$\textbf{Ours-P}$}&50.4&$\textbf{67.5}$&$\textbf{87.8}$&$\textbf{92.1}$&$\textbf{42.2}$&$\textbf{68.0}$\\ 
			\bottomrule 
		\end{tabular}
\end{adjustbox}
	
\end{table*}

\subsubsection{AU Detection in Source Domain}
We compare our AU detection results in source domain to the state-of-the-art advanced self-supervised approaches, including six typical self-supervised AU detection methods $\textbf{SLTC}$~\citep{lu:et.al:2020:BMVC}, $\textbf{Fab-Net}$~\citep{wiles:et.al:2018:BMVC-a}, $\textbf{TAE}$~\citep{li:et.al:2022:TPAMI}, $\textbf{EmoCo}$~\citep{sun:et.al:2021:FG}, $\textbf{ADLD}$~\citep{shao:et.al:2022:TAC}, $\textbf{KSRL}$~\citep{chang:et.al:2022:CVPR}, and five representative contrastive representation learning methods $\textbf{SimCLR}$~\citep{chen:et.al:2020:ICML}, $\textbf{SimSiam}$~\citep{chen:et.al:2021:CVPR}, $\textbf{MoCo}$~\citep{he:et.al:2020:CVPR}, $\textbf{CVC}$~\citep{wu:et.al:2021:ICCV}, $\textbf{CLP}$~\citep{li:et.al:2023:TIP} in Table.~\ref{tab:4} and Table.~\ref{tab:5}. Although these methods are all based on unsupervised self-supervised frameworks, but they all used AU labels from source domain datasets to verify the experimental results.

\begin{table*}[tbp] 
	\centering
\small
		\caption{F1-score of our source domain results and state-of-the-art methods on BP4D, where the target domain dataset is EmotioNet.}
		\label{tab:4}
		\tabcolsep=0.4cm
		\begin{adjustbox}{max width=\linewidth}
\begin{tabular}{l c c c c c c c}
			\toprule
			\multicolumn{1}{l}{\multirow{2}{*}{$\textbf{Method}$}}&\multicolumn{6}{c}{$\textbf{AU Index}$}&\multicolumn{1}{c}{\multirow{2}{*}{$\textbf{Avg.}$}}\\ 
			\cline{2-7}
			\multicolumn{1}{l}{}&$\textbf{1}$&$\textbf{2}$&$\textbf{4}$&$\textbf{6}$&$\textbf{12}$&$\textbf{17}$& \\
			\hline  
			\multicolumn{1}{l}{$\textbf{SimCLR}$~\citep{chen:et.al:2020:ICML}}&38.0&36.4&37.2&66.6&76.2&56.1&51.8\\
			\multicolumn{1}{l}{$\textbf{SLTC}$~\citep{lu:et.al:2020:BMVC}}&42.3&24.3&44.1&71.8&83.3&51.6&52.9\\
			\multicolumn{1}{l}{$\textbf{SimSiam}$~\citep{chen:et.al:2021:CVPR}}&38.6&39.0&37.2&68.2&77.6&57.9&53.1\\
			\multicolumn{1}{l}{$\textbf{Fab-Net}$~\citep{wiles:et.al:2018:BMVC-a}}&43.3&35.7&41.6&72.9&83.5&48.2&54.2\\
			\multicolumn{1}{l}{$\textbf{MoCo}$~\citep{he:et.al:2020:CVPR}}&30.8&41.3&42.1&70.2&82.5&59.1&54.3\\
			\multicolumn{1}{l}{$\textbf{CVC}$~\citep{wu:et.al:2021:ICCV}}&43.9&47.8&38.7&67.0&84.4&49.3&55.2\\
			\multicolumn{1}{l}{$\textbf{EmoCo}$~\citep{sun:et.al:2021:FG}}&45.4&30.5&55.5&76.1&$\textbf{87.6}$&59.1&59.0\\
			\multicolumn{1}{l}{$\textbf{ADLD}$~\citep{shao:et.al:2022:TAC}}&50.5&35.7&$\textbf{61.8}$&74.1&75.2&69.0&61.0\\
			\multicolumn{1}{l}{$\textbf{TAE}$~\citep{li:et.al:2022:TPAMI}}&47.0&45.9&50.9&74.7&85.6&62.3&61.1\\
			\multicolumn{1}{l}{$\textbf{CLP}$~\citep{li:et.al:2023:TIP}}&47.7&$\textbf{50.9}$&49.5&75.8&84.1&62.7&61.8\\
			\multicolumn{1}{l}{$\textbf{KSRL}$~\citep{chang:et.al:2022:CVPR}}&50.1&45.4&53.6&$\textbf{79.2}$&87.4&56.1&62.0\\
			\multicolumn{1}{l}{$\textbf{Ours}$}&$\textbf{54.0}$&38.2&61.5&78.5&77.5&$\textbf{72.0}$&$\textbf{63.6}$\\ 
			\bottomrule 
		\end{tabular}
\end{adjustbox}
	
\end{table*}

\begin{table*}[tbp] 
	\centering
\small
		\caption{F1-score of our source domain results and state-of-the-art methods on DISFA, where the target domain dataset is EmotioNet.}
		\label{tab:5}
		\tabcolsep=0.4cm
		\begin{adjustbox}{max width=\linewidth}
\begin{tabular}{l c c c c c c}
			\toprule
			\multicolumn{1}{l}{\multirow{2}{*}{$\textbf{Method}$}}&\multicolumn{5}{c}{$\textbf{AU Index}$}&\multicolumn{1}{c}{\multirow{2}{*}{$\textbf{Avg.}$}}\\ 
			\cline{2-6}
			\multicolumn{1}{l}{}&$\textbf{4}$&$\textbf{6}$&$\textbf{12}$&$\textbf{25}$&$\textbf{26}$& \\
			\hline  
			\multicolumn{1}{l}{$\textbf{SLTC}$~\citep{lu:et.al:2020:BMVC}}&35.1&33.6&67.5&68.0&43.8&49.6\\
			\multicolumn{1}{l}{$\textbf{Fab-Net}$~\citep{wiles:et.al:2018:BMVC-a}}&43.2&50.4&69.6&72.4&42.4&55.6\\
			\multicolumn{1}{l}{$\textbf{SimCLR}$~\citep{chen:et.al:2020:ICML}}&47.5&42.4&66.8&81.5&52.7&58.2\\
			\multicolumn{1}{l}{$\textbf{MoCo}$~\citep{he:et.al:2020:CVPR}}&45.9&45.4&72.9&83.4&54.5&60.2\\
			\multicolumn{1}{l}{$\textbf{CVC}$~\citep{wu:et.al:2021:ICCV}}&56.4&49.5&75.5&79.1&55.7&64.2\\
			\multicolumn{1}{l}{$\textbf{EmoCo}$~\citep{sun:et.al:2021:FG}}&63.9&52.5&77.0&78.3&44.2&63.2\\
			\multicolumn{1}{l}{$\textbf{SimSiam}$~\citep{chen:et.al:2021:CVPR}}&58.1&$\textbf{53.8}$&74.4&79.0&55.7&64.2\\
			\multicolumn{1}{l}{$\textbf{TAE}$~\citep{li:et.al:2022:TPAMI}}&64.5&46.8&73.2&85.1&55.3&65.0\\
			\multicolumn{1}{l}{$\textbf{KSRL}$~\citep{chang:et.al:2022:CVPR}}&58.1&52.5&$\textbf{77.6}$&86.9&53.2&65.7\\
			\multicolumn{1}{l}{$\textbf{Ours}$}&$\textbf{65.7}$&40.2&74.6&$\textbf{89.0}$&$\textbf{61.1}$&$\textbf{66.1}$\\ 
			\bottomrule 
		\end{tabular}
\end{adjustbox}
	
\end{table*}

The results on BP4D are shown in Table.~\ref{tab:4}. Although our source domain AU detection network structure is relatively simple, our method achieves an average increase of 10.7\%, 9.4\%, 4.6\%, 2.5\% to SLTC, Fab-Net, EmoCo, TAE. This is because these methods only concentrate on global facial features, overlooking task-related domain knowledge such as the distinctive properties of AUs. Compared to contrastive representation learning methods, namely SimCLR, SimSiam, MoCo, CVC and CLP, our method outperforms by around 1.8\%-11.8\% in F1-score. This occurs because our approach takes full advantage of static datasets, while the prevalent use of self-supervised tasks based on random crops or temporal consistency in contrastive approaches can lead to insufficient AU representations. The reason for the 2.6\% improvement over ADLD is that loading pre-training parameters help the initial facial extractor $E_{f}$($I_{s}$, $F_{s}$) warm up before training. Compared to the second best approach KSRL, our method has demonstrated superior performance by increasing the F1-score from 62.0\% to 63.6\%. It should be noted that compared with KSRL training in laboratory scene, using backgrounds in the wild greatly improves the generalization performance of our framework. 

Similar results can be observed on DISFA as summarized in Table.~\ref{tab:5}, where our method achieves the best overall AU detection performance.

\subsubsection{Comparision with facial landmark detection methods}
\begin{figure}[tbp] 
	\centering
	\includegraphics[width=0.99\linewidth]{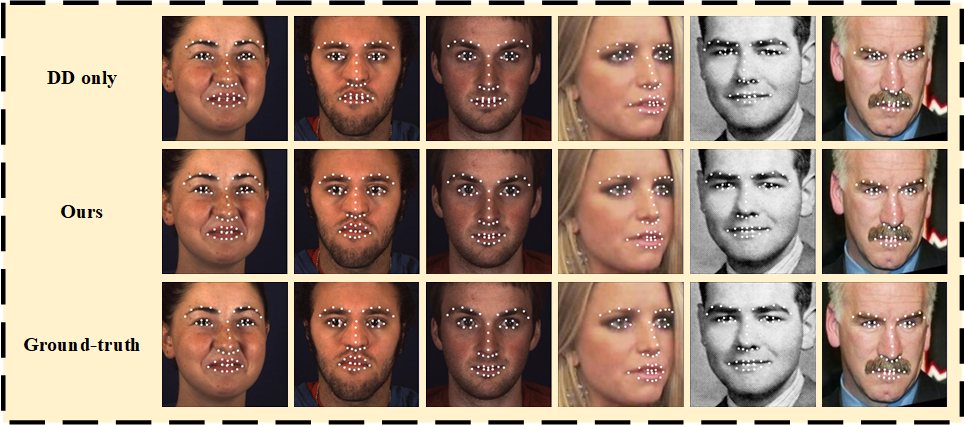}
	\caption{Qualitative results of facial landmark detection in both source domain and target domain.}
	\label{fig:4.png}
\end{figure}

Due to the different ways of obtaining the ground truths of facial landmark detection from target domain dataset EmotioNet, there is no certain standard for evaluating the accuracy facial landmark detection in target domain at present. Table.~\ref{tab:6} only shows the mean error and failure rate results of our facial landmark detection framework on BP4D. Additionally, we compare our method with state-of-the-art facial landmark detection methods, including $\textbf{MCL}$~\citep{shao:et.al:2020:Neurocomputing}, $\textbf{TCDCN}$~\citep{zhang:et.al:2015:TPAMI}, $\textbf{ERT}$~\citep{kazemi:et.al:2014:CVPR}, $\textbf{LAB}$~\citep{wu:et.al:2018:CVPR}, $\textbf{HRNetV2}$~\citep{wang:et.al:2020:TPAMI}, $\textbf{OpenPose}$~\citep{he:et.al:2017:FG}, $\textbf{J$\hat{\rm \textbf{A}}$A-Net}$~\citep{shao:et.al:2021:IJCV}. The qualitative results of facial landmark detection in both source domain and target domain are shown in Fig.~\ref{fig:4.png}.

\begin{table}[tbp] 
	\centering
\small
		\caption{Mean error (lower is better) and failure rate (lower is better) results of different facial landmark detection methods on BP4D. }
		\label{tab:6}
		\tabcolsep=0.05cm
		\begin{adjustbox}{max width=\linewidth}
\begin{tabular}{l c c}
			\toprule
			\multicolumn{1}{l}{$\textbf{Method}$}&$\textbf{Mean Error}$&$\textbf{Failure Rate}$\\
			\hline  
			\multicolumn{1}{l}{$\textbf{MCL}$~\citep{shao:et.al:2020:Neurocomputing}}&7.20&1.69\\
			\multicolumn{1}{l}{$\textbf{TCDCN}$~\citep{zhang:et.al:2015:TPAMI}}&6.57&1.88\\
			\multicolumn{1}{l}{$\textbf{ERT}$~\citep{kazemi:et.al:2014:CVPR}}&4.73&3.49\\
			\multicolumn{1}{l}{$\textbf{LAB}$~\citep{wu:et.al:2018:CVPR}}&4.50&0.11\\
			\multicolumn{1}{l}{$\textbf{HRNetV2}$~\citep{wang:et.al:2020:TPAMI}}&4.35&$\textbf{0.02}$\\
			\multicolumn{1}{l}{$\textbf{OpenPose}$~\citep{he:et.al:2017:FG}}&3.93&0.27\\
			\multicolumn{1}{l}{$\textbf{J$\hat{\rm \textbf{A}}$A-Net}$~\citep{shao:et.al:2021:IJCV}}&3.80&0.32\\
			\multicolumn{1}{l}{$\textbf{Ours(DD only)}$}&5.87&1.16\\
			\multicolumn{1}{l}{$\textbf{Ours}$}&$\textbf{3.16}$&0.19\\
			\bottomrule 
		\end{tabular}
\end{adjustbox}
	
\end{table}

It can be found that our methods exhibit superior performance compared to MCL, TCDCN, ERT, OpenPose, and J$\hat{\rm A}$A-Net, excelling in both mean error and failure rate metrics. Notably, despite being specifically tailored for AU detection, our approach outperforms even recently proposed methods such as LAB and HRNetV2 in terms of mean error, showcasing the lowest mean error overall.

Besides, with the help of AU domain separation and reconstruction task and AU detection task, our method outperforms dual domains facial landmark detection framework (DD) training only with large margins of 2.71\% and 0.97\% for mean error and failure rate respectively. This demonstrates that other facial tasks are also beneficial for facial landmark detection. These three facial tasks contribute to each other in our proposed multi-task learning strategy.

\subsection{Ablation Study}
Quantitative results of our framework with different component combinations are summarized in Table.~\ref{tab:7}, including the baseline (BL), the source domain facial landmark detection (SD), the target domain facial landmark detection (TD), the pixel-level feature alignment scheme (AS), the projectors for alignment (AP) and weighted asymmetric loss (AL).

\begin{table*}[tbp] 
	\centering
\small
		\caption{F1-scores on EmotioNet when the source domain is BP4D for ablation study. }
		\label{tab:7}
		\tabcolsep=0.3cm
		\begin{adjustbox}{max width=\linewidth}
\begin{tabular}{c c c c c c c c c}
			\toprule
			\multicolumn{1}{c}{$\textbf{Model}$}&BL&SD&TD&AS&AP&AL&$\textbf{F1-Score}$&$\textbf{Accuracy}$\\
			\hline  
			\multicolumn{1}{c}{A}&$\checkmark$&&&&&&36.5&86.4\\
			\multicolumn{1}{c}{B}&$\checkmark$&$\checkmark$&&&&&37.5(+1.0)&87.1(+0.7)\\
			\multicolumn{1}{c}{C}&$\checkmark$&&$\checkmark$&&&&37.8(+1.3)&87.3(+0.9)\\
			\multicolumn{1}{c}{D}&$\checkmark$&$\checkmark$&$\checkmark$&&&&38.6(+2.1)&87.6(+1.2)\\
			\multicolumn{1}{c}{E}&$\checkmark$&$\checkmark$&$\checkmark$&$\checkmark$&&&39.0(+2.5)&87.7(+1.3)\\
			\multicolumn{1}{c}{F}&$\checkmark$&$\checkmark$&$\checkmark$&$\checkmark$&$\checkmark$&&40.2(+3.7)&88.1(+1.7)\\
			\multicolumn{1}{c}{G}&$\checkmark$&$\checkmark$&$\checkmark$&$\checkmark$&$\checkmark$&$\checkmark$&$\textbf{42.1(+5.6)}$&$\textbf{88.6(+2.2)}$\\
			\bottomrule 
		\end{tabular}
\end{adjustbox}
	
\end{table*}

$\textbf{Contribution of multi-task learning strategy:}$ As summarized in Table.~\ref{tab:7}, the introduction of dual domains facial landmark detection framework helps achieve an increase of 2.1\% in F1-score and 1.2\% in Accuracy comparing to the baseline, validating that weight sharing of homostructural facial extraction modules can promote the training accuracy of two similar facial tasks. At the same time, the results of two facial landmark detection branches training together exceed the results of one of the branches training alone, which also demonstrate the effectiveness of multi-task learning strategy.

$\textbf{Integrating of pixel-level feature alignment scheme:}$ The introduction of the pixel-level feature alignment scheme without projectors helps achieve an average increase of 0.4\% in F1-score comparing to model D, which illustrates that adding four intermediate supervisors can play a certain role in better feature reconstruction. Moreover, the result of model F sharply increased to 40.2\% by adding projectors before reconstructed features, since projectors with simple structures add the number of neural network layers, thus increasing the fitting ability of our framework.

$\textbf{Effectiveness of weighted asymmetric loss:}$ Our proposed weighted asymmetric loss leads to an increase of 1.9\% in F1-score comparing to model F by changing the contribution of positive and negative samples of each type of AUs to model parameters updating. This demonstrates that the training process concentrates more on the small number of samples, making the distribution of positive samples have a more significant impact on the binary classification predicting results. 

\begin{figure*}[tbp]
	\centering
	\includegraphics[width=0.99\linewidth]{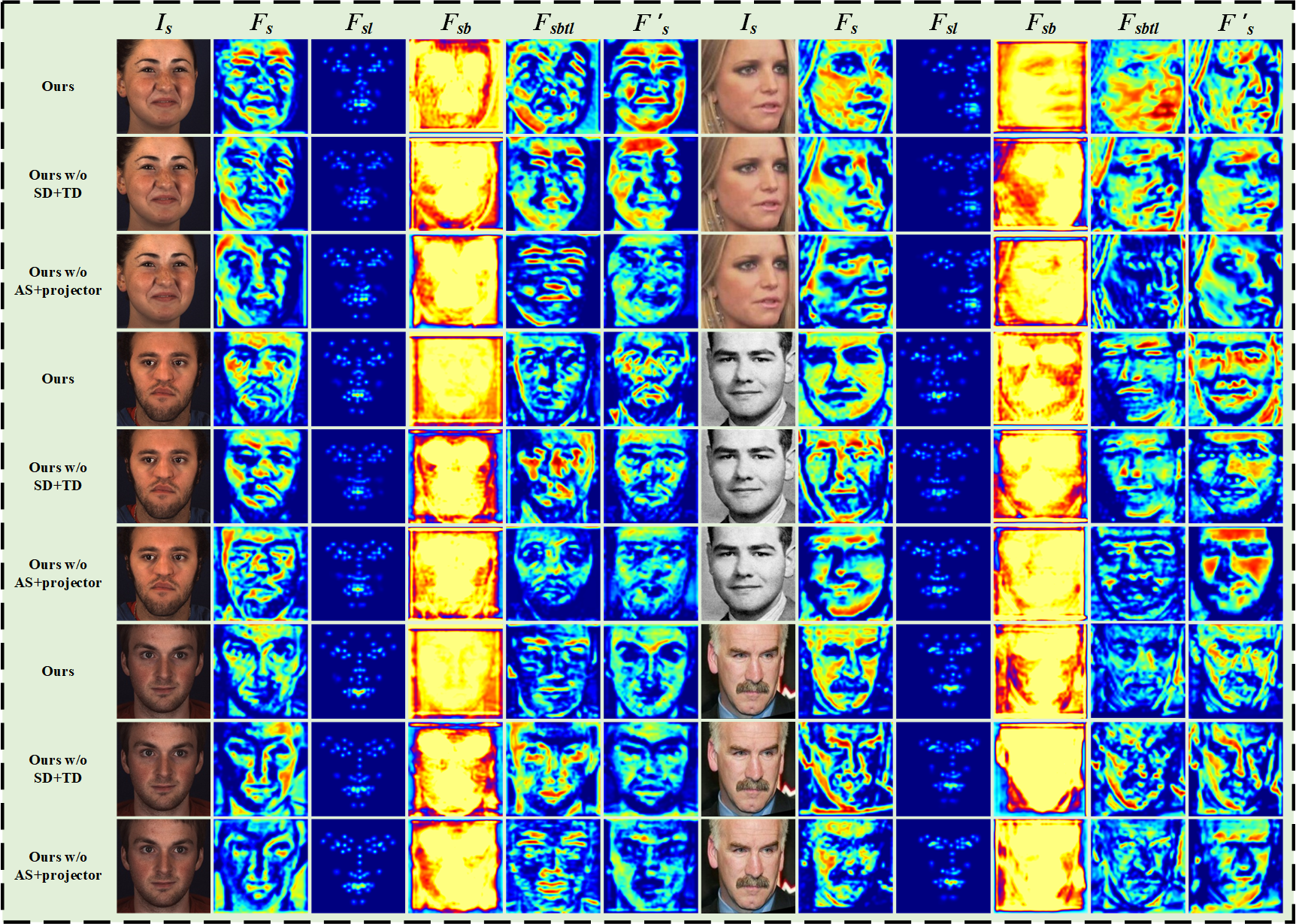}
	\caption{Qualitative results of the ablation studies in both source domain and target domain.}
	\label{fig:5.png}
\end{figure*}

Based on ablation studies, introducing each component is beneficial for performance improvement, among which AL is the most effective component. Qualitative results of the ablation studies are shown in Fig.~\ref{fig:5.png}.
\subsection{Hyperparameters Analysis} \label{Hyperparameters Analysis}
\subsubsection{\texorpdfstring{${\lambda}_{(.)}$}{Loss weights} in the overall loss \texorpdfstring{$L_{all}$}{L-all}} 
\begin{figure*}[tbp]
	\centering
	\includegraphics[width=0.3\textwidth]{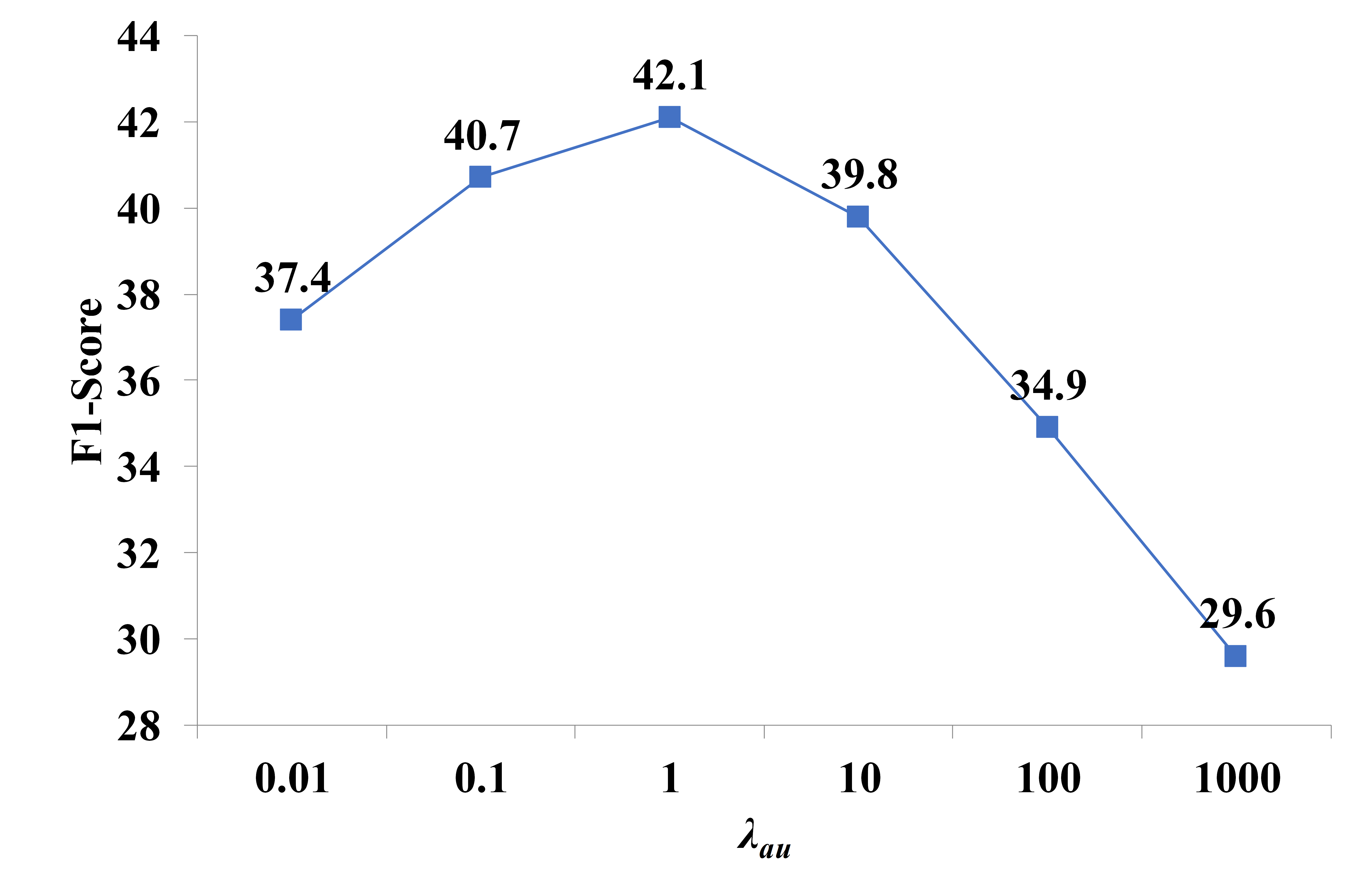}
	\includegraphics[width=0.3\textwidth]{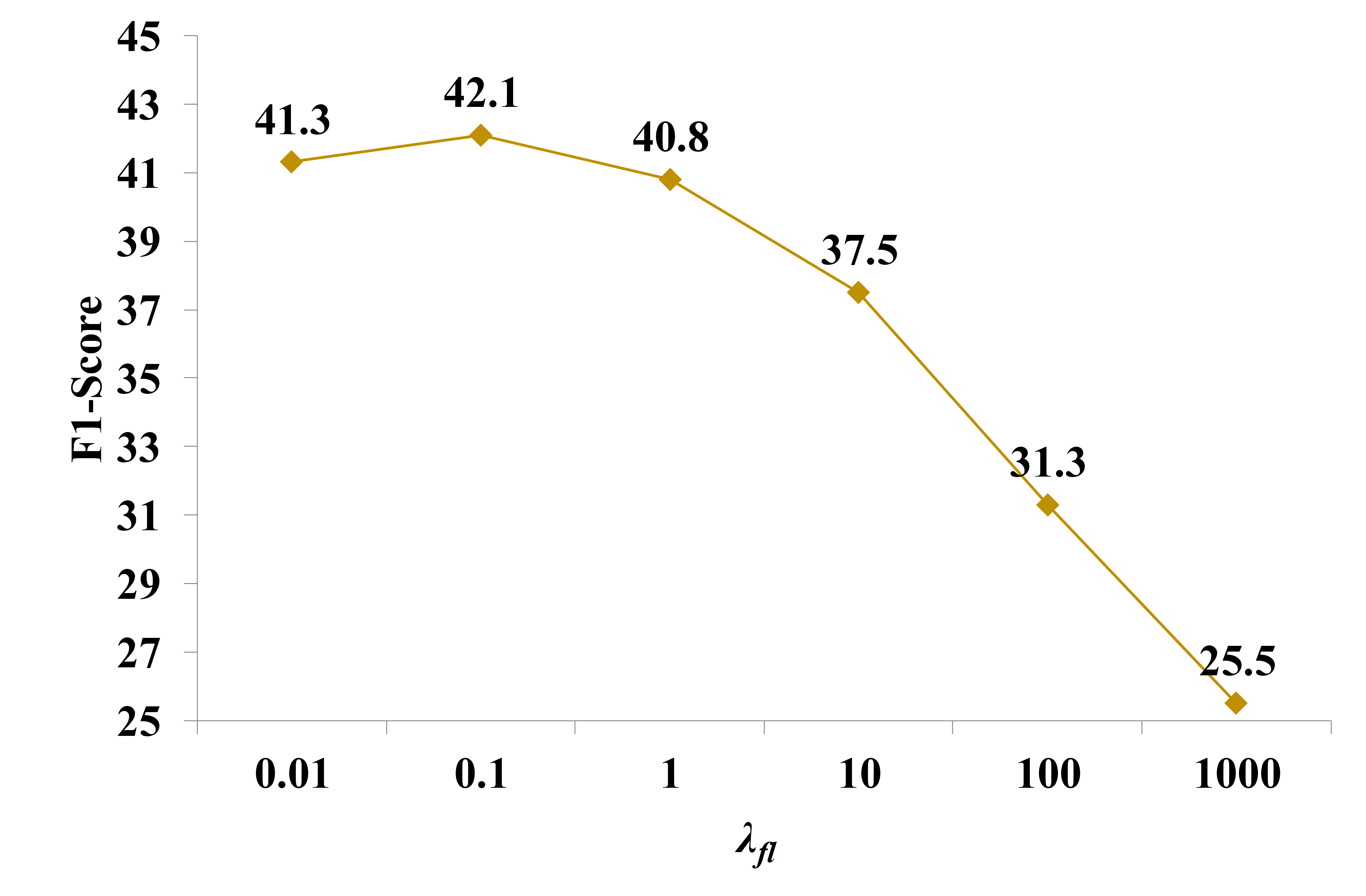}
	\includegraphics[width=0.3\textwidth]{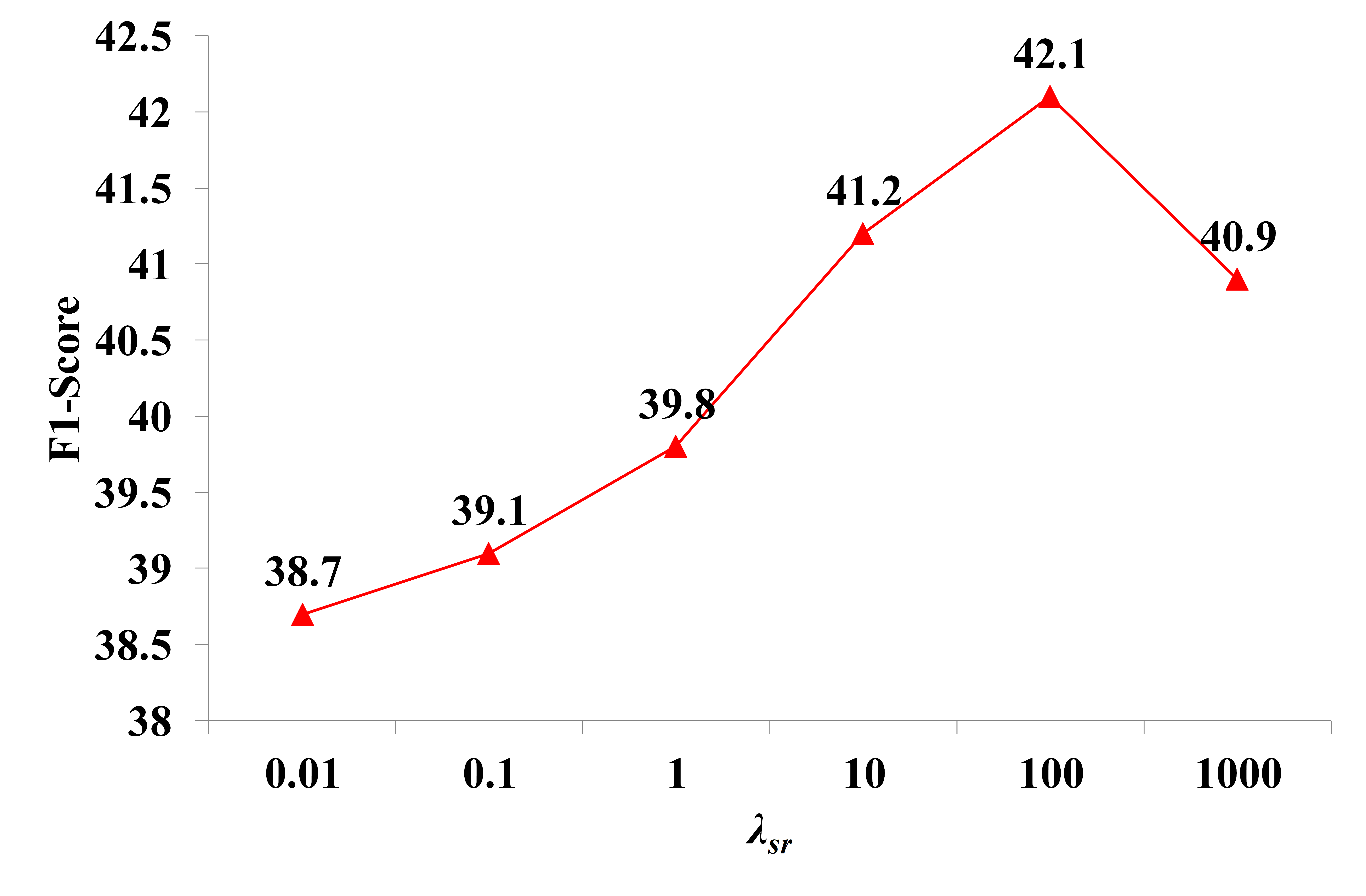}
	\caption{The analysis of hyperparameter ${\lambda}_{(.)}$ in the overall loss $L_{all}$.}
	\label{fig:6.png}
\end{figure*}

Due to $\lambda_{lr}$, $\lambda_{adlb}$, $\lambda_{adst}$ are fixed to ensure the separation effect, we only vary $\lambda_{au}$, $\lambda_{fl}$ and $\lambda_{sr}$ from 0.01 to 1000, to investigate its influence on our framework. The experiment results are respectively shown in Fig.~\ref{fig:6.png}. We observe that $\lambda_{sr}$ works best at 100, which illustrates that the improved self-reconstruction loss plays a very important role in the training process. The reason why $\lambda_{fl}$ should not be large is to avoid convergence of features of two different facial tasks.
\subsubsection{Focusing parameter \texorpdfstring{$\beta$}{beta}} 
\begin{figure}[tbp] 
	\centering
	\includegraphics[width=0.8\linewidth]{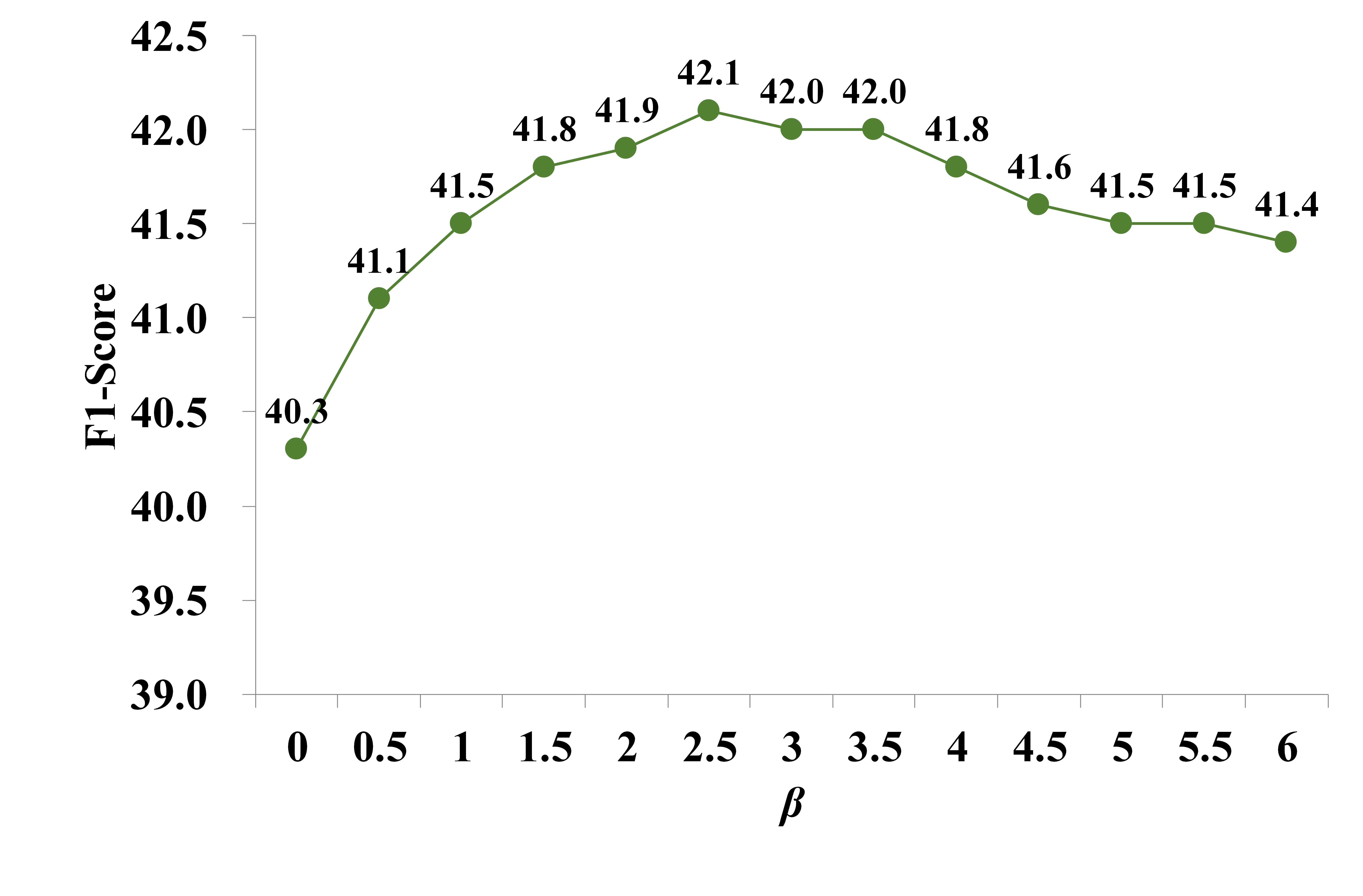}
	\caption{The analysis of hyperparameter $\beta$.}
	\label{fig:7.png}
\end{figure}

As we know, the focusing parameter $\beta$ plays a important role in mitigating the impact of easily classified negative samples. To systematically explore this influence, we systematically vary the hyperparameter $\beta$ over the range from 0 to 6.0, while maintaining $\gamma$ at a constant value of 0.3. The resulting changes in experimental outcomes are visually presented in Fig.~\ref{fig:7.png}. Our observations indicate that the optimal performance is achieved when $\beta$ is set at 2.5. This demonstrates that the loss fails to impart a sufficient degree of down-weighting for easily classified negative samples when $\beta$ is below this threshold. Conversely, when $\beta$ surpasses this threshold, an excessive down-weighting of the infrequent positive samples occurs.

\subsubsection{Asymmetric probability margin \texorpdfstring{$\gamma$}{gamma}} 
\begin{figure}[tbp] 
	\centering
	\includegraphics[width=0.8\linewidth]{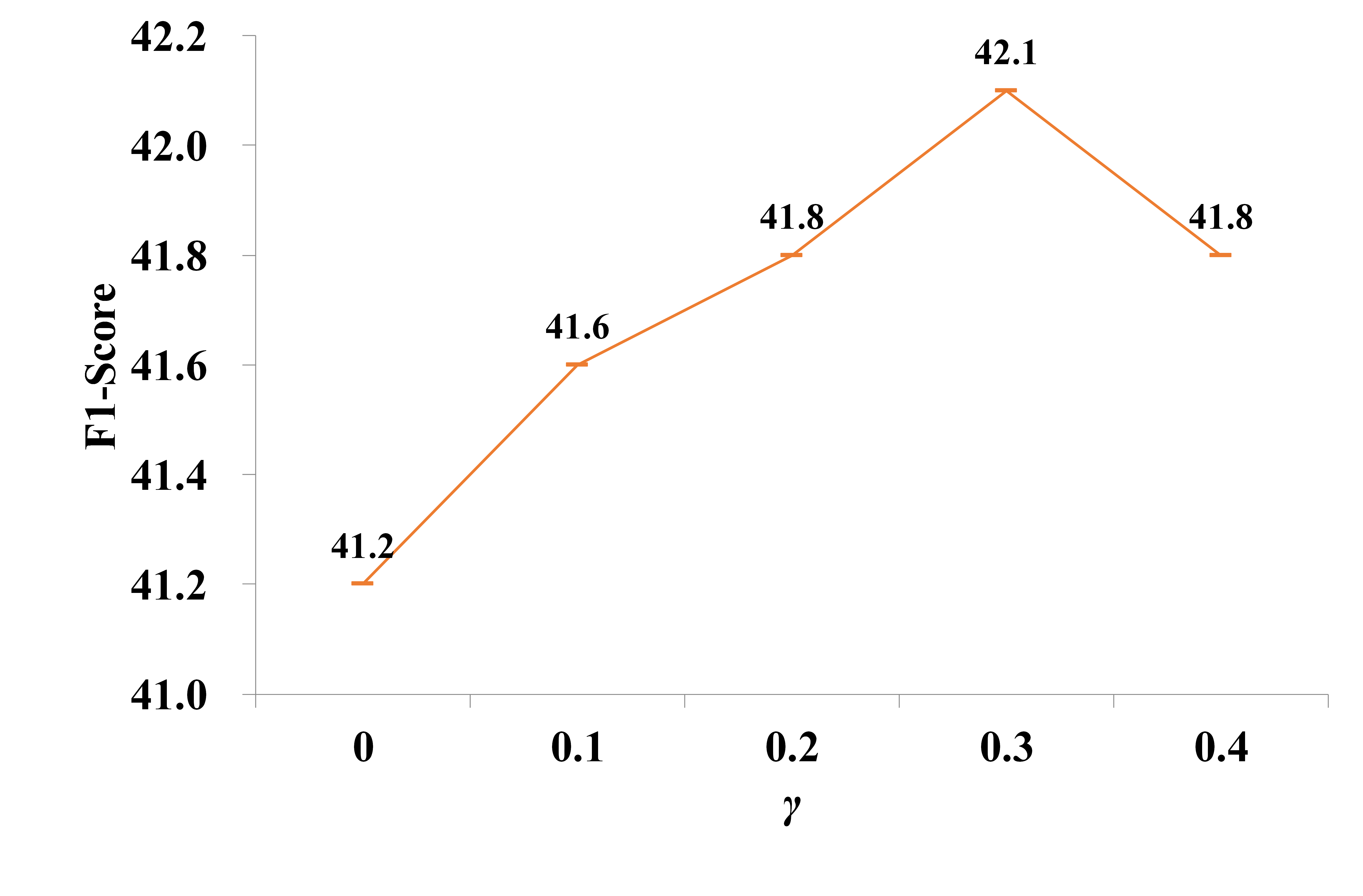}
	\caption{The analysis of hyperparameter $\gamma$.}
	\label{fig:8.png}
\end{figure}

As illustrated in Fig.~\ref{fig:8.png}, we vary our asymmetric probability margin $\gamma=\{0, 0.1, 0.2, 0.3, 0.4\}$ maintaining a fixed focusing parameter at $\beta$ = 2.5. The results reveal that the optimal probability margin is attained at $\gamma$ = 0.3, yielding a notable F1-score of 42.1\%. This optimal configuration is attributed to the non-linear attenuation effects induced by the focusing parameter $\beta$, wherein $\gamma$=0.3 proves particularly effective in facilitating the ignoration of very easy nagetive samples and the rejection of mislabeled nagetive samples.

\FloatBarrier
\section{Conclusion}
In this paper, we propose a multi-task learning strategy boosting AU detection in the wild through jointing facial landmark detection and AU domain separation and reconstruction. Two facial landmark detection branches help solve the lack of accurate facial landmark coordinates during the AU domain separation and reconstruction training process, while the parameters of homostructural facial extraction modules from these two tasks are shared. Apart from that, our proposed pixel-level feature alignment scheme maintains the consistency of features obtained from two separation and reconstruction processes. In addition, the weighted asymmetric loss respectively changes the contribution of positive and negative samples of each type of AUs to model parameters updating. Experimental results demonstrate the superior performance of our framework for AU detection in the wild. 
\FloatBarrier
\section*{Acknowledgement}
The computation is completed in the HPC Platform of Huazhong University of Science and Technology.

\FloatBarrier
\bibliographystyle{elsarticle-harv}
\bibliography{references}

@book{rosenberg:et.al:2020,
	title={What the face reveals: Basic and applied studies of spontaneous expression using the Facial Action Coding System (FACS)},
	author={Rosenberg, Erika L and Ekman, Paul},
	year={2020},
	publisher={Oxford University Press}
}

@article{dahmane:et.al:2014:TMM,
	title={Prototype-based modeling for facial expression analysis},
	author={Dahmane, Mohamed and Meunier, Jean},
	journal={IEEE Transactions on Multimedia},
	volume={16},
	number={6},
	pages={1574--1584},
	year={2014},
	publisher={IEEE}
}

@inproceedings{zhao:et.al:2016:CVPR,
	title={Deep region and multi-label learning for facial action unit detection},
	author={Zhao, Kaili and Chu, Wen-Sheng and Zhang, Honggang},
	booktitle={Proceedings of the IEEE conference on computer vision and pattern recognition (CVPR)},
	pages={3391--3399},
	year={2016}
}

@inproceedings{corneanu:et.al:2018:ECCV,
	title={Deep structure inference network for facial action unit recognition},
	author={Corneanu, Ciprian and Madadi, Meysam and Escalera, Sergio},
	booktitle={Proceedings of the european conference on computer vision (ECCV)},
	pages={298--313},
	year={2018}
}

@article{li:et.al:2018:TPAMI,
	title={Eac-net: Deep nets with enhancing and cropping for facial action unit detection},
	author={Li, Wei and Abtahi, Farnaz and Zhu, Zhigang and Yin, Lijun},
	journal={IEEE transactions on pattern analysis and machine intelligence},
	volume={40},
	number={11},
	pages={2583--2596},
	year={2018},
	publisher={IEEE}
}

@article{ma:et.al:2019:Neurocomputing,
	title={Au r-cnn: Encoding expert prior knowledge into r-cnn for action unit detection},
	author={Ma, Chen and Chen, Li and Yong, Junhai},
	journal={Neurocomputing},
	volume={355},
	pages={35--47},
	year={2019},
	publisher={Elsevier}
}

@article{bauer:et.al:2008:IJHR,
	title={Human--robot collaboration: a survey},
	author={Bauer, Andrea and Wollherr, Dirk and Buss, Martin},
	journal={International Journal of Humanoid Robotics},
	volume={5},
	number={01},
	pages={47--66},
	year={2008},
	publisher={World Scientific}
}

@inproceedings{mcduff:et.al:2013:FG,
	title={Predicting online media effectiveness based on smile responses gathered over the internet},
	author={McDuff, Daniel and El Kaliouby, Rana and Demirdjian, David and Picard, Rosalind},
	booktitle={2013 10th IEEE international conference and workshops on automatic face and gesture recognition (FG)},
	pages={1--7},
	year={2013},
	organization={IEEE}
}

@inproceedings{szirtes:et.al:2013:ER,
	title={Facing reality: An industrial view on large scale use of facial expression analysis},
	author={Szirtes, G{\'a}bor and Szolgay, D{\'a}niel and Utasi, {\'A}kos and Tak{\'a}cs, D{\'a}niel and Petr{\'a}s, Istv{\'a}n and Fodor, G{\'a}bor},
	booktitle={Proceedings of the 2013 on Emotion recognition in the wild challenge and workshop},
	pages={1--8},
	year={2013}
}

@article{vicente:et.al:2015:TITS,
	title={Driver gaze tracking and eyes off the road detection system},
	author={Vicente, Francisco and Huang, Zehua and Xiong, Xuehan and De la Torre, Fernando and Zhang, Wende and Levi, Dan},
	journal={IEEE Transactions on Intelligent Transportation Systems},
	volume={16},
	number={4},
	pages={2014--2027},
	year={2015},
	publisher={IEEE}
}

@inproceedings{wiles:et.al:2018:BMVC-a,
	title={Self-supervised learning of a facial attribute embedding from video},
	author={Wiles, Olivia and Koepke, A and Zisserman, Andrew},
	booktitle={British Machine Vision Conference (BMVC)},
	pages={1--15},
	year={2018},
	publisher={IEEE}
}

@inproceedings{lu:et.al:2020:BMVC,
	title={Self-Supervised Learning for Facial Action Unit Recognition through Temporal Consistency},
	author={Lu, Liupei and Tavabi, Leili and Soleymani, Mohammad},
	booktitle={British Machine Vision Conference (BMVC)},
	pages={1--8},
	year={2020},
	publisher={IEEE}
}

@inproceedings{sun:et.al:2021:FG,
	title={Emotion-aware contrastive learning for facial action unit detection},
	author={Sun, Xuran and Zeng, Jiabei and Shan, Shiguang},
	booktitle={2021 16th IEEE International Conference on Automatic Face and Gesture Recognition (FG)},
	pages={1--8},
	year={2021},
	publisher={IEEE}
}

@article{li:et.al:2022:TPAMI,
	title={Learning representations for facial actions from unlabeled videos},
	author={Li, Yong and Zeng, Jiabei and Shan, Shiguang},
	journal={IEEE Transactions on Pattern Analysis and Machine Intelligence},
	volume={44},
	number={1},
	pages={302--317},
	year={2022},
	publisher={IEEE}
}

@article{li:et.al:2023:TIP,
	title={Contrastive Learning of Person-independent Representations for Facial Action Unit Detection},
	author={Li, Yong and Shan, Shiguang},
	journal={IEEE Transactions on Image Processing},
	year={2023},
	publisher={IEEE}
}

@article{wang:et.al:2017:arXiv,
	title={Transferring face verification nets to pain and expression regression},
	author={Wang, Feng and Xiang, Xiang and Liu, Chang and Tran, Trac D and Reiter, Austin and Hager, Gregory D and Quon, Harry and Cheng, Jian and Yuille, Alan L},
	journal={arXiv preprint arXiv:1702.06925},
	year={2017}
}

@inproceedings{zhao:et.al:2018:CVPR,
	title={Learning facial action units from web images with scalable weakly supervised clustering},
	author={Zhao, Kaili and Chu, Wen-Sheng and Martinez, Aleix M},
	booktitle={Proceedings of the IEEE Conference on computer vision and pattern recognition (CVPR)},
	pages={2090--2099},
	year={2018}
}

@article{niu:et.al:2019:NIPS,
	title={Multi-label co-regularization for semi-supervised facial action unit recognition},
	author={Niu, Xuesong and Han, Hu and Shan, Shiguang and Chen, Xilin},
	journal={Advances in neural information processing systems},
	volume={32},
	year={2019}
}

@article{yan:et.al:2023:TMM,
	title={Weakly supervised regional and temporal learning for facial action unit recognition},
	author={Yan, Jingwei and Wang, Jingjing and Li, Qiang and Wang, Chunmao and Pu, Shiliang},
	journal={IEEE Transactions on Multimedia},
	year={2023},
	publisher={IEEE}
}

@inproceedings{yang:et.al:2023:ICONIP,
	title={Landmark-Assisted Facial Action Unit Detection with Optimal Attention and Contrastive Learning},
	author={Yang, Yi and Hu, Qiaoping and Lu, Hongtao and Jiang, Fei and Li, Yaoyi},
	booktitle={International Conference on Neural Information Processing (ICONIP)},
	pages={92--104},
	year={2023},
	organization={Springer}
}

@inproceedings{liu:et.al:2023:ICASSP,
	title={Semi-Supervised Contrastive Learning with Soft Mask Attention for Facial Action Unit Detection},
	author={Liu, Zhongling and Liu, Rujie and Shi, Ziqiang and Liu, Liu and Mi, Xiaoyu and Murase, Kentaro},
	booktitle={IEEE International Conference on Acoustics, Speech and Signal Processing (ICASSP)},
	pages={1--5},
	year={2023},
	organization={IEEE}
}

@inproceedings{wiles:et.al:2018:BMVC-b,
	title={X2face: A network for controlling face generation using images, audio, and pose codes},
	author={Wiles, Olivia and Koepke, A and Zisserman, Andrew},
	booktitle={Proceedings of the European conference on computer vision (ECCV)},
	pages={670--686},
	year={2018}
}

@inproceedings{li:et.al:2019:CVPR,
	title={Self-supervised representation learning from videos for facial action unit detection},
	author={Li, Yong and Zeng, Jiabei and Shan, Shiguang and Chen, Xilin},
	booktitle={Proceedings of the IEEE/CVF Conference on Computer vision and pattern recognition (CVPR)},
	pages={10924--10933},
	year={2019}
}

@article{ganin:et.al:2016,
	title={Domain-adversarial training of neural networks},
	author={Ganin, Yaroslav and Ustinova, Evgeniya and Ajakan, Hana and Germain, Pascal and Larochelle, Hugo and Laviolette, Fran{\c{c}}ois and March, Mario and Lempitsky, Victor},
	journal={Journal of machine learning research},
	volume={17},
	number={59},
	pages={1--35},
	year={2016}
}

@inproceedings{tzeng:et.al:2017:CVPR,
	title={Adversarial discriminative domain adaptation},
	author={Tzeng, Eric and Hoffman, Judy and Saenko, Kate and Darrell, Trevor},
	booktitle={Proceedings of the IEEE conference on computer vision and pattern recognition (CVPR)},
	pages={7167--7176},
	year={2017}
}

@inproceedings{lee:et.al:2018:ECCV,
	title={Diverse image-to-image translation via disentangled representations},
	author={Lee, Hsin-Ying and Tseng, Hung-Yu and Huang, Jia-Bin and Singh, Maneesh and Yang, Ming-Hsuan},
	booktitle={Proceedings of the European conference on computer vision (ECCV)},
	pages={35--51},
	year={2018}
}

@inproceedings{zheng:et.al:2018:ECCV,
	title={T2net: Synthetic-to-realistic translation for solving single-image depth estimation tasks},
	author={Zheng, Chuanxia and Cham, Tat-Jen and Cai, Jianfei},
	booktitle={Proceedings of the European conference on computer vision (ECCV)},
	pages={767--783},
	year={2018}
}

@article{goodfellow:et.al:2014:generative,
	title={Generative adversarial nets},
	author={Goodfellow, Ian and Pouget-Abadie, Jean and Mirza, Mehdi and Xu, Bing and Warde-Farley, David and Ozair, Sherjil and Courville, Aaron and Bengio, Yoshua},
	journal={Advances in neural information processing systems},
	volume={27},
	year={2014}
}

@inproceedings{wang:et.al:2018:ACM,
	title={Personalized multiple facial action unit recognition through generative adversarial recognition network},
	author={Wang, Can and Wang, Shangfei},
	booktitle={Proceedings of the 26th ACM international conference on Multimedia (ACM MM)},
	pages={302--310},
	year={2018}
}

@article{shao:et.al:2022:TAC,
	title={Unconstrained facial action unit detection via latent feature domain},
	author={Shao, Zhiwen and Cai, Jianfei and Cham, Tat-Jen and Lu, Xuequan and Ma, Lizhuang},
	journal={IEEE Transactions on Affective Computing},
	volume={13},
	number={2},
	pages={1111--1126},
	year={2022},
	publisher={IEEE}
}

@inproceedings{jyoti:et.al:2018:DICTA,
	title={A single hierarchical network for face, action unit and emotion detection},
	author={Jyoti, Shreyank and Sharma, Garima and Dhall, Abhinav},
	booktitle={2018 Digital Image Computing: Techniques and Applications (DICTA)},
	pages={1--8},
	year={2018},
	organization={IEEE}
}

@article{tallec:et.al:2022:TAC,
	title={Multi-order networks for action unit detection},
	author={Tallec, Gauthier and Dapogny, Arnaud and Bailly, Kevin},
	journal={IEEE Transactions on Affective Computing},
	year={2022},
	publisher={IEEE}
}

@article{shao:et.al:2021:IJCV,
	title={Jaa-net: joint facial action unit detection and face alignment via adaptive attention},
	author={Shao, Zhiwen and Liu, Zhilei and Cai, Jianfei and Ma, Lizhuang},
	journal={International Journal of Computer Vision},
	volume={129},
	pages={321--340},
	year={2021},
	publisher={Springer}
}

@article{wang:et.al:2023:TAC,
	title={Dual Learning for Joint Facial Landmark Detection and Action Unit Recognition},
	author={Wang, Shangfei and Chang, Yanan and Wang, Can},
	journal={IEEE Transactions on Affective Computing},
	year={2023},
	publisher={IEEE}
}

@article{king:et.al:2012:access,
	title={Dlib c++ library},
	author={King, Davis},
	journal={Access on: http://dlib. net},
	year={2012}
}

@inproceedings{zhu:et.al:2016:CVPR,
	title={Face alignment across large poses: A 3d solution},
	author={Zhu, Xiangyu and Lei, Zhen and Liu, Xiaoming and Shi, Hailin and Li, Stan Z},
	booktitle={Proceedings of the IEEE conference on computer vision and pattern recognition (CVPR)},
	pages={146--155},
	year={2016}
}

@inproceedings{simon:et.al:2017:CVPR,
	title={Hand keypoint detection in single images using multiview bootstrapping},
	author={Simon, Tomas and Joo, Hanbyul and Matthews, Iain and Sheikh, Yaser},
	booktitle={Proceedings of the IEEE conference on Computer Vision and Pattern Recognition (CVPR)},
	pages={1145--1153},
	year={2017}
}

@inproceedings{wu:et.al:2018:CVPR,
	title={Look at boundary: A boundary-aware face alignment algorithm},
	author={Wu, Wayne and Qian, Chen and Yang, Shuo and Wang, Quan and Cai, Yici and Zhou, Qiang},
	booktitle={Proceedings of the IEEE conference on computer vision and pattern recognition (CVPR)},
	pages={2129--2138},
	year={2018}
}

@inproceedings{bhattacharya:et.al:2020:ECCV,
	title={Take an emotion walk: Perceiving emotions from gaits using hierarchical attention pooling and affective mapping},
	author={Bhattacharya, Uttaran and Roncal, Christian and Mittal, Trisha and Chandra, Rohan and Kapsaskis, Kyra and Gray, Kurt and Bera, Aniket and Manocha, Dinesh},
	booktitle={Proceedings of the European conference on computer vision (ECCV)},
	pages={145--163},
	year={2020},
	organization={Springer}
}

@inproceedings{li:et.al:2017:CVPR,
	title={Action unit detection with region adaptation, multi-labeling learning and optimal temporal fusing},
	author={Li, Wei and Abtahi, Farnaz and Zhu, Zhigang},
	booktitle={Proceedings of the IEEE conference on computer vision and pattern recognition (CVPR)},
	pages={1841--1850},
	year={2017}
}

@article{zhou:et.al:2022:IJMLC,
	title={Learning deep representation for action unit detection with auxiliary facial attributes},
	author={Zhou, Caixia and Zhi, Ruicong},
	journal={International Journal of Machine Learning and Cybernetics},
	pages={1--13},
	year={2022},
	publisher={Springer}
}

@inproceedings{chen:et.al:2020:ICML,
	title={A simple framework for contrastive learning of visual representations},
	author={Chen, Ting and Kornblith, Simon and Norouzi, Mohammad and Hinton, Geoffrey},
	booktitle={International conference on machine learning (ICML)},
	pages={1597--1607},
	year={2020},
	organization={PMLR}
}

@inproceedings{he:et.al:2020:CVPR,
	title={Momentum contrast for unsupervised visual representation learning},
	author={He, Kaiming and Fan, Haoqi and Wu, Yuxin and Xie, Saining and Girshick, Ross},
	booktitle={Proceedings of the IEEE/CVF conference on computer vision and pattern recognition (CVPR)},
	pages={9729--9738},
	year={2020}
}

@inproceedings{chen:et.al:2021:CVPR,
	title={Exploring simple siamese representation learning},
	author={Chen, Xinlei and He, Kaiming},
	booktitle={Proceedings of the IEEE/CVF conference on computer vision and pattern recognition (CVPR)},
	pages={15750--15758},
	year={2021}
}

@inproceedings{yan:et.al:2021:ACM,
	title={Self-supervised regional and temporal auxiliary tasks for facial action unit recognition},
	author={Yan, Jingwei and Wang, Jingjing and Li, Qiang and Wang, Chunmao and Pu, Shiliang},
	booktitle={Proceedings of the 29th ACM International Conference on Multimedia (ACM MM)},
	pages={1038--1046},
	year={2021}
}

@inproceedings{zhang:et.al:2019:CVPR,
	title={Context-aware feature and label fusion for facial action unit intensity estimation with partially labeled data},
	author={Zhang, Yong and Jiang, Haiyong and Wu, Baoyuan and Fan, Yanbo and Ji, Qiang},
	booktitle={Proceedings of the IEEE/CVF International Conference on Computer Vision (CVPR)},
	pages={733--742},
	year={2019}
}

@article{cui:et.al:2020,
	title={Knowledge augmented deep neural networks for joint facial expression and action unit recognition},
	author={Cui, Zijun and Song, Tengfei and Wang, Yuru and Ji, Qiang},
	journal={Advances in Neural Information Processing Systems},
	volume={33},
	pages={14338--14349},
	year={2020}
}

@inproceedings{chang:et.al:2022:CVPR,
	title={Knowledge-driven self-supervised representation learning for facial action unit recognition},
	author={Chang, Yanan and Wang, Shangfei},
	booktitle={Proceedings of the IEEE/CVF Conference on Computer Vision and Pattern Recognition (CVPR)},
	pages={20417--20426},
	year={2022}
}

@article{zhang:et.al:2015:TPAMI,
	title={Learning deep representation for face alignment with auxiliary attributes},
	author={Zhang, Zhanpeng and Luo, Ping and Loy, Chen Change and Tang, Xiaoou},
	journal={IEEE transactions on pattern analysis and machine intelligence},
	volume={38},
	number={5},
	pages={918--930},
	year={2015},
	publisher={IEEE}
}

@article{shao:et.al:2020:Neurocomputing,
	title={Deep multi-center learning for face alignment},
	author={Shao, Zhiwen and Zhu, Hengliang and Tan, Xin and Hao, Yangyang and Ma, Lizhuang},
	journal={Neurocomputing},
	volume={396},
	pages={477--486},
	year={2020},
	publisher={Elsevier}
}

@inproceedings{honari:et.al:2016:CVPR,
	title={Recombinator networks: Learning coarse-to-fine feature aggregation},
	author={Honari, Sina and Yosinski, Jason and Vincent, Pascal and Pal, Christopher},
	booktitle={Proceedings of the IEEE/CVF Conference on Computer Vision and Pattern Recognition (CVPR)},
	pages={5743--5752},
	year={2016}
}

@inproceedings{xiao:et.al:2016:ECCV,
	title={Robust facial landmark detection via recurrent attentive-refinement networks},
	author={Xiao, Shengtao and Feng, Jiashi and Xing, Junliang and Lai, Hanjiang and Yan, Shuicheng and Kassim, Ashraf},
	booktitle={Proceedings of the European conference on computer vision (ECCV)},
	pages={57--72},
	year={2016},
	organization={Springer}
}

@inproceedings{woo:et.al:2018:ECCV,
	title={CBAM: Convolutional block attention module},
	author={Woo, Sanghyun and Park, Jongchan and Lee, Joon-Young and Kweon, In So},
	booktitle={Proceedings of the European conference on computer vision (ECCV)},
	pages={3--19},
	year={2018}
}

@inproceedings{chen:et.al:2022:knowledge,
	title={Knowledge distillation with the reused teacher classifier},
	author={Chen, Defang and Mei, Jian-Ping and Zhang, Hailin and Wang, Can and Feng, Yan and Chen, Chun},
	booktitle={Proceedings of the IEEE/CVF Conference on Computer Vision and Pattern Recognition (CVPR)},
	pages={11933--11942},
	year={2022}
}

@article{adriana:et.al:2015:fitnets,
	title={Fitnets: Hints for thin deep nets},
	author={Adriana, Romero and Nicolas, Ballas and Ebrahimi, K Samira and Antoine, Chassang and Carlo, Gatta and Yoshua, B},
	journal={International Conference on Learning Representations (ICLR)},
	volume={2},
	pages={3},
	year={2015}
}

@inproceedings{yang:et.al:2020:knowledge,
	title={Knowledge distillation via softmax regression representation learning},
	author={Yang, Jing and Martinez, Brais and Bulat, Adrian and Tzimiropoulos, Georgios},
	booktitle={International Conference on Learning Representations (ICLR)},
	year={2020}
}

@article{shang:et.al:2023:PRL,
	title={MMA-Net: Multi-view Mixed Attention Mechanism for Facial Action Unit Detection},
	author={Shang, Ziqiao and Du, Congju and Li, Bingyin and Yan, Zengqiang and Yu, Li},
	journal={Pattern Recognition Letters},
	year={2023},
	publisher={Elsevier}
}

@inproceedings{ridnik:et.al:2021:ICCV,
	title={Asymmetric loss for multi-label classification},
	author={Ridnik, Tal and Ben-Baruch, Emanuel and Zamir, Nadav and Noy, Asaf and Friedman, Itamar and Protter, Matan and Zelnik-Manor, Lihi},
	booktitle={Proceedings of the IEEE/CVF International Conference on Computer Vision (ICCV)},
	pages={82--91},
	year={2021}
}

@article{luo:et.al:2022:IJCAI,
	title={Learning multi-dimensional edge feature-based au relation graph for facial action unit recognition},
	author={Luo, Cheng and Song, Siyang and Xie, Weicheng and Shen, Linlin and Gunes, Hatice},
	booktitle={Proceedings of the International Joint Conference on Artificial Intelligence (IJCAI)},
	year={2022}
}

@article{zhang:et.al:2014:IVC,
	title={Bp4d-spontaneous: a high-resolution spontaneous 3d dynamic facial expression database},
	author={Zhang, Xing and Yin, Lijun and Cohn, Jeffrey F and Canavan, Shaun and Reale, Michael and Horowitz, Andy and Liu, Peng and Girard, Jeffrey M},
	journal={Image and Vision Computing},
	volume={32},
	number={10},
	pages={692--706},
	year={2014},
	publisher={Elsevier}
}

@article{mavadati:et.al:2013:TAC,
	title={Disfa: A spontaneous facial action intensity database},
	author={Mavadati, S Mohammad and Mahoor, Mohammad H and Bartlett, Kevin and Trinh, Philip and Cohn, Jeffrey F},
	journal={IEEE Transactions on Affective Computing},
	volume={4},
	number={2},
	pages={151--160},
	year={2013},
	publisher={IEEE}
}

@inproceedings{fabian:et.al:2016:CVPR,
	title={Emotionet: An accurate, real-time algorithm for the automatic annotation of a million facial expressions in the wild},
	author={Fabian Benitez-Quiroz, C and Srinivasan, Ramprakash and Martinez, Aleix M},
	booktitle={Proceedings of the IEEE/CVF Conference on Computer Vision and Pattern Recognition (CVPR)},
	pages={5562--5570},
	year={2016}
}

@inproceedings{xiong:et.al:2013:CVPR,
	title={Supervised descent method and its applications to face alignment},
	author={Xiong, Xuehan and De la Torre, Fernando},
	booktitle={Proceedings of the IEEE/CVF Conference on Computer Vision and Pattern Recognition (CVPR)},
	pages={532--539},
	year={2013}
}

@article{cootes:et.al:2001:TPAMI,
	title={Active appearance models},
	author={Cootes, Timothy F. and Edwards, Gareth J. and Taylor, Christopher J},
	journal={IEEE Transactions on pattern analysis and machine intelligence},
	volume={23},
	number={6},
	pages={681--685},
	year={2001},
	publisher={IEEE}
}

@inproceedings{kinga:et.al:2015:ICLR,
	title={A method for stochastic optimization},
	author={Kinga, D and Adam, Jimmy Ba and others},
	booktitle={International conference on learning representations (ICLR)},
	volume={5},
	pages={6},
	year={2015},
	organization={San Diego, California;}
}

@article{shao:et.al:2019:TAC,
	title={Facial action unit detection using attention and relation learning},
	author={Shao, Zhiwen and Liu, Zhilei and Cai, Jianfei and Wu, Yunsheng and Ma, Lizhuang},
	journal={IEEE transactions on affective computing},
	volume={13},
	number={3},
	pages={1274--1289},
	year={2019},
	publisher={IEEE}
}

@inproceedings{kazemi:et.al:2014:CVPR,
	title={One millisecond face alignment with an ensemble of regression trees},
	author={Kazemi, Vahid and Sullivan, Josephine},
	booktitle={Proceedings of the IEEE conference on computer vision and pattern recognition (CVPR)},
	pages={1867--1874},
	year={2014}
}

@article{wang:et.al:2020:TPAMI,
	title={Deep high-resolution representation learning for visual recognition},
	author={Wang, Jingdong and Sun, Ke and Cheng, Tianheng and Jiang, Borui and Deng, Chaorui and Zhao, Yang and Liu, Dong and Mu, Yadong and Tan, Mingkui and Wang, Xinggang and others},
	journal={IEEE transactions on pattern analysis and machine intelligence},
	volume={43},
	number={10},
	pages={3349--3364},
	year={2020},
	publisher={IEEE}
}

@inproceedings{he:et.al:2017:FG,
	title={Multi view facial action unit detection based on CNN and BLSTM-RNN},
	author={He, Jun and Li, Dongliang and Yang, Bin and Cao, Siming and Sun, Bo and Yu, Lejun},
	booktitle={2017 12th IEEE International Conference on Automatic Face \& Gesture Recognition (FG)},
	pages={848--853},
	year={2017},
	organization={IEEE}
}

@inproceedings{wu:et.al:2021:ICCV,
	title={Contrastive learning of image representations with cross-video cycle-consistency},
	author={Wu, Haiping and Wang, Xiaolong},
	booktitle={Proceedings of the IEEE/CVF International Conference on Computer Vision (ICCV)},
	pages={10149--10159},
	year={2021}
}

@inproceedings{baltrusaitis:et.al:2018:FG,
	title={Openface 2.0: Facial behavior analysis toolkit},
	author={Baltrusaitis, Tadas and Zadeh, Amir and Lim, Yao Chong and Morency, Louis-Philippe},
	booktitle={2018 13th IEEE international conference on automatic face \& gesture recognition (FG)},
	pages={59--66},
	year={2018},
	organization={IEEE}
}

@article{wan:et.al:2021:TIP,
	title={Robust face alignment by multi-order high-precision hourglass network},
	author={Wan, Jun and Lai, Zhihui and Liu, Jun and Zhou, Jie and Gao, Can},
	journal={IEEE Transactions on Image Processing},
	volume={30},
	pages={121--133},
	year={2021},
	publisher={IEEE}
}

@article{wan:et.al:2022:TNNLS,
	title={Robust facial landmark detection by multiorder multiconstraint deep networks},
	author={Wan, Jun and Lai, Zhihui and Li, Jing and Zhou, Jie and Gao, Can},
	journal={IEEE Transactions on Neural Networks and Learning Systems},
	volume={33},
	number={5},
	pages={2181--2194},
	year={2022},
	publisher={IEEE}
}

@inproceedings{yang:et.al:2017:CVPR,
	title={Stacked hourglass network for robust facial landmark localisation},
	author={Yang, Jing and Liu, Qingshan and Zhang, Kaihua},
	booktitle={Proceedings of the IEEE/CVF conference on computer vision and pattern recognition workshops (CVPR workshop)},
	pages={79--87},
	year={2017}
}

@inproceedings{dong:et.al:2018:CVPR,
	title={Style aggregated network for facial landmark detection},
	author={Dong, Xuanyi and Yan, Yan and Ouyang, Wanli and Yang, Yi},
	booktitle={Proceedings of the IEEE/CVF conference on computer vision and pattern recognition (CVPR)},
	pages={379--388},
	year={2018}
}

@inproceedings{liu:et.al:2019:CVPR,
	title={Semantic alignment: Finding semantically consistent ground-truth for facial landmark detection},
	author={Liu, Zhiwei and Zhu, Xiangyu and Hu, Guosheng and Guo, Haiyun and Tang, Ming and Lei, Zhen and Robertson, Neil M and Wang, Jinqiao},
	booktitle={Proceedings of the IEEE/CVF conference on computer vision and pattern recognition (CVPR)},
	pages={3467--3476},
	year={2019}
}

@article{wan:et.al:2023:TIP,
	title={Precise Facial Landmark Detection by Reference Heatmap Transformer},
	author={Wan, Jun and Liu, Jun and Zhou, Jie and Lai, Zhihui and Shen, Linlin and Sun, Hang and Xiong, Ping and Min, Wenwen},
	journal={IEEE Transactions on Image Processing},
	volume={32},
	pages={1966--1977},
	year={2023},
	publisher={IEEE}
}
\end{document}